\documentclass[journal]{IEEEtran}
\ifCLASSINFOpdf
\else
\fi
\usepackage[tight,footnotesize]{subfigure}

\usepackage{times}
\usepackage{epsfig}
\usepackage{graphicx}
\usepackage{amsmath}
\usepackage{amssymb}
\usepackage{epstopdf}

\usepackage{algorithm}
\usepackage{algpseudocode}
\usepackage{color}
\usepackage{multirow}
\usepackage{setspace}

\usepackage[
  none
]{hyphenat}
\usepackage{theorem}

\newcommand{\set}[1]{
  \left\{ #1 \right\}
}
\newcommand{\floor}[1]{
  \left\lfloor #1 \right\rfloor
}
\newcommand{\prob}[1]{
  P\left\{ #1 \right\}
}

\DeclareMathOperator{\tree}{tree}

\usepackage[pagebackref=true,breaklinks=false,colorlinks,bookmarks=false]{hyperref}

\begin{document}
%
\title{Retrieval in Long Surveillance Videos using User Described Motion \& Object Attributes}
%
%
%

\author{Greg~Casta{\~n\'o}n,
			  Mohamed~Elgharib,
        Venkatesh~Saligrama,
        and~Pierre-Marc~Jodoin \vspace{-0.7cm}}
%
%

\markboth{IEEE Transactions on Multimedia,~Vol.~XX, No.~XXX, JanXX~20XX}%
{Shell \MakeLowercase{\textit{et al.}}: Bare Demo of IEEEtran.cls for Journals}
%



\maketitle

\begin{abstract}

We present a content-based retrieval method for long surveillance videos both for wide-area (Airborne) as well as near-field imagery (CCTV). Our goal is to retrieve video segments, with a focus on detecting objects moving on routes, that match user-defined events of interest.  The sheer size and remote locations where surveillance videos are acquired, necessitates highly compressed representations that are also meaningful for supporting user-defined queries. To address these challenges we archive long-surveillance video through lightweight processing based on low-level local spatio-temporal extraction of motion and object features. These are then hashed into an inverted index using locality-sensitive hashing (LSH). This local approach allows for query flexibility as well as leads to significant gains in compression. Our second task is to extract partial matches to the user-created query and assembles them into full matches using Dynamic Programming (DP). DP exploits causality to assemble the indexed low level features into a video segment which matches the query route. We examine CCTV and Airborne footage, whose low contrast makes motion extraction more difficult. We generate robust motion estimates for Airborne data using a tracklets generation algorithm while we use Horn and Schunck approach to generate motion estimates for CCTV. Our approach handles long routes, low contrasts and occlusion. We derive bounds on the rate of false positives and demonstrate the effectiveness of the approach for counting, motion pattern recognition and abandoned object applications. 

\end{abstract}


%
\IEEEpeerreviewmaketitle

\IEEEpeerreviewmaketitle
\vspace{-0.4cm}
\section{Introduction}
\label{sec:introduction}
\vspace{-0.1cm}

Video surveillance camera networks are increasingly ubiquitous, generating thousands of hours of archived video every day. This data is rarely processed in real-time and primarily used for scene investigation purposes to gather evidence after events take place. In military applications UAVs produce terabytes of wide area imagery in real-time at remote/hostile locations. Both of these cases necessitate maintaining highly compressed searchable representations that are local to the user but yet sufficiently informative and flexible to handle a wide range of queries. While compression is in general lossy from the perspective of video reconstruction it is actually desirable from the perspective of search since it not only reduces the search space but it also leverages the fact that for a specific query most data is irrelevant and pre-processing procedures such as video summarization is often unnecessary and inefficient. Consequently, we need techniques that are memory and run-time efficient to scale with large data sets, and be able to retrieve video segments matching user defined queries with robustness to common problems, such as low frame-rate, low contrast and occlusion. 



  Some of the main challenges of this problem are:  
  
  \noindent
  {\bf 1.) Data lifetime}: since video is constantly streamed, there is a perpetual
  renewal of video data.  This calls for a model that can be updated
  incrementally as video data is made available.  The model must also scale 
  with the temporal mass of the video.

  \noindent
  {\bf 2.) Unpredictable queries}: the nature of queries depends on the field of
  view of the camera, the scene itself and the type of events being observed.
  The system should support queries of different nature that can retrieve both
  recurrent events such as people entering a store and infrequent events such as
  abandoned objects and cars performing U-turns.

  \noindent
  {\bf 3.) Unpredictable event duration}: events are unstructured. They start
  anytime, vary in length, and overlap with other events.  The system is
  nonetheless expected to return complete events regardless of their duration
  and whether or not other events occur simultaneously.

  \noindent
  {\bf 4.) Clutter and occlusions}: Tracking and tagging objects in urban videos is
  challenging due to occlusions and clutter; especially when real-time
  performance is required.

  This paper presents a method for fast content-based retrieval adapted to characteristics of surveillance videos.  Our technique is query-driven where the user draws routes manually. The algorithm then retrieves video segments containing objects that moved in the user supplied routes. Our technique consists of two steps: The first extracts low level features and hashes them into an inverted index using locality-sensitive hashing (LSH). The second stage extracts partials matches and assembles them into full matches using Dynamic Programming (DP). 
  
\begin{figure}
    \centering
		\subfigure[]
		{
      \includegraphics[width=.6\linewidth]{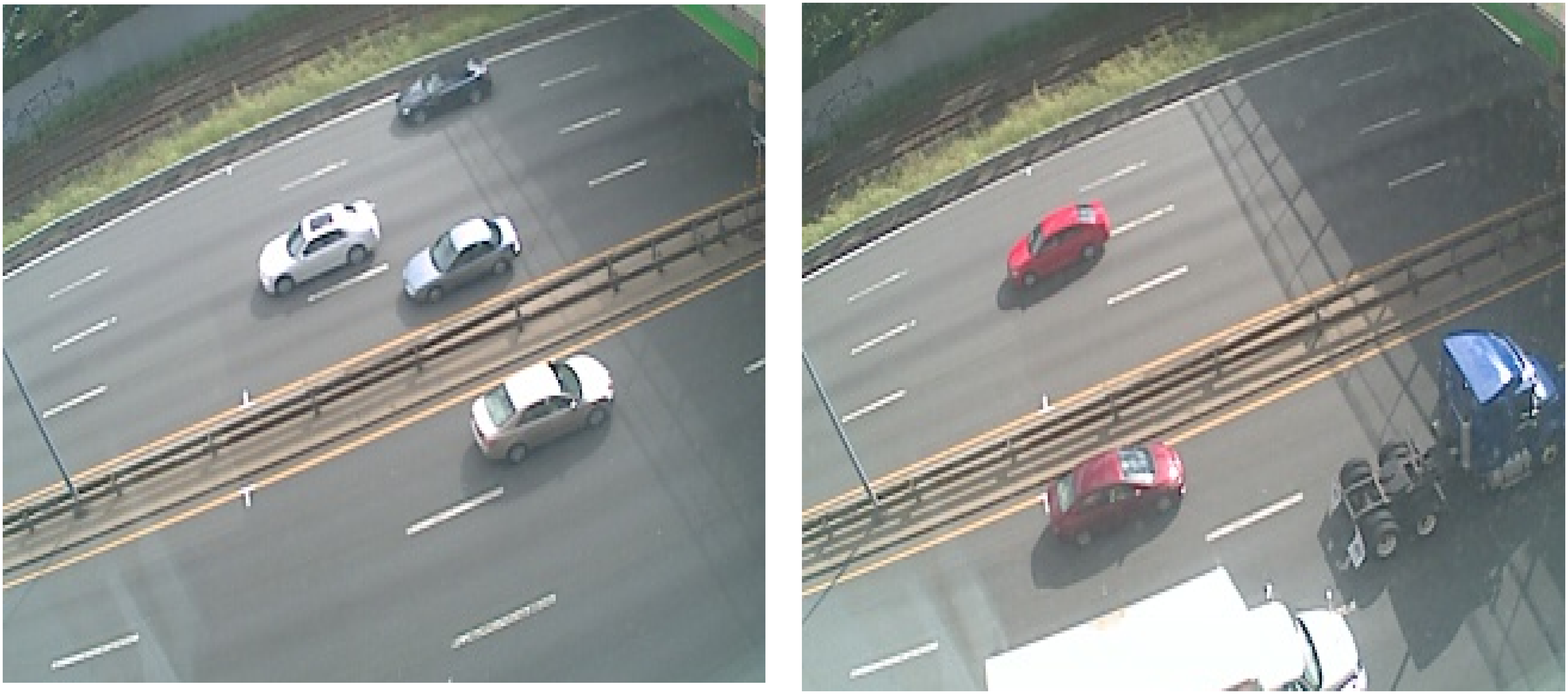}
		}
		\subfigure[]
		{
			\includegraphics[width=.998\linewidth]{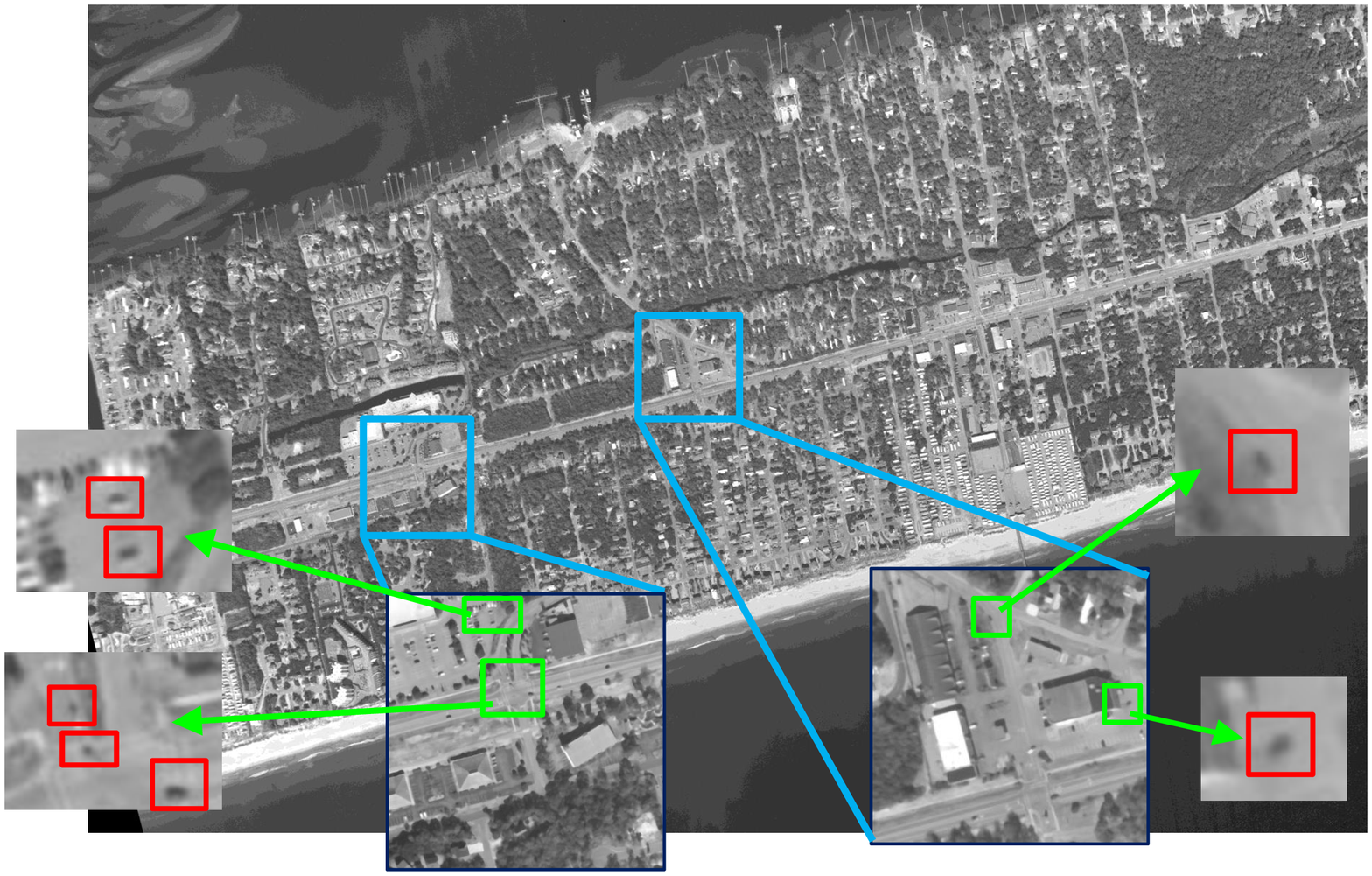}
    }
    \caption{\small Top: Examples of CCTV footage. Bottom: Example of Airborne footage. Blue and green boxes show zoomed-in regions. Objects of interests (cars in this figure) usually have much lower contrast in Airborne footage (see red boxes) over CCTV footage. This makes processing Airborne footage more challenging than CCTV.\vspace{-0.3cm}}
    \label{fig:DataIllustration}
	\end{figure}

  We examine two types of data: (1) CCTV and (2) Airborne. Airborne footage contains aerial view images shot from a UAV. Their processing is more challenging than CCTV due to their lower frame rate and lower contrast (see Fig.~\ref{fig:DataIllustration} ). From this footage we are able to extract a number of low level features, described in Section \ref{sec:feature-extraction}. Here certain features such as optical flow information (extracted using Horn and Schunck technique \cite{Horn81}) are rendered highly noisy due to the Airborne low contrast. Hence for Airborne footage we first generate tracklets, using ideas from \cite{pitie05,baugh09}, and then generate motion estimates from those tracklets. Our technique retrieves routes generated by different objects (such as cars and human), handles long routes, low-contrast, occlusion, and performs faster than current techniques.


  %
  %

\vspace{-0.2cm}
\section{Related Work}
\label{s:related-work}
  
 Most video papers devoted to summarization and search focus on broadcast videos such as music clips, sports games, movies, etc.  These methods typically divide the video into  ``shots"~\cite{Zhu05,Sujatha11} by locating key frames corresponding to scene transitions.  The search procedure exploits key frame content and matches either low-level descriptors~\cite{Sujatha11} or higher-level semantic meta-tags to a given query~\cite{Zhu05}.

Surveillance videos are fundamentally different than conventional videos.  Not only do surveillance videos have no global motion induced by a moving camera, they often show unrelated or unimportant moving and static objects.  For that reason, surveillance videos are difficult to decompose into ``scenes" separated by key frames that one could summarize with some meta tags or a global mathematical model.  Furthermore, surveillance videos generally have no closed-caption or audio track ~\cite{Zhu05}.

For that reason, search in a surveillance video focuses on understanding and indexing the dynamic content of the video in a way that is compatible with arbitrary upcoming user-defined queries.  Thus, most scene-understanding video analytic methods work on a two-stage procedure: 
\begin{itemize}
\item Learn patterns of activities via a clustering/learning procedure 
\item Recognize new patterns of activity via a classification stage.  
\end{itemize}
For the first stage many methods are based on the fact that activities in public areas often follow some basic rules (traffic lights, highways, building entries, etc). Therefore, during the learning stage these methods often quantize space and time into a number of states with transition probabilities.  Common models are HMMs~\cite{Malinici08,Kuettel10,Jouneau11}, Bayesian networks~\cite{Calderara07,Xiang08}, context free grammars~\cite{Veeraraghavan07}, and other graphical models~\cite{Simon09,Wang09}.  During the classification stage these learned patterns are either used to recognize pre-defined patterns of activity~\cite{Yilmaz08,Gorelick07,Simon09,Shechtman07, Yeo08} (useful for counting~\cite{Tian08}) or to detect anomalies by flagging everything that deviates from what has been previously learned~\cite{Saligrama10,Malinici08,Jouneau11}.  We also note that that some of these methods that account for global behavior understanding often rely on tracking~\cite{Jouneau11,Malinici08,Calderara07} while those devoted to isolated action recognition rely significantly on low-level features~\cite{Gorelick07,Shechtman07,Yilmaz08}.

Although these methods could probably be adapted to index the video and facilitate search, very few methods this problem explicitly~\cite{Little13}.  An exception is a method based on topic modeling by Wang {\em et al.}~\cite{Wang09}.  Their method decomposes the video into {\em clips} in which the local motion is quantized into {\em words}.  These words are than clustered into so-called {\em topics} such that each clip is modeled as a distribution over these topics.  They consider queries that are mixtures of these topics. Consequently, their search algorithm fetches every clip containing all of the topics mentioned in the query.  A similar approach can be found in \cite{Yang09,Emonet14,Xiang08, Kuettel10}
But search techniques focused on global explanations operate at a competitive disadvantage: the preponderance of clutter (4th issue in the introduction) in surveillance video makes the training step of scene understanding prohibitively difficult.  Second, since these techniques often focus on understanding recurrent activities, they are unsuited for retrieving infrequent events - this can be a problem, given that queries are unpredictable (2nd issue in the introduction), and some activities of interest may seldom occur.  Finally, the training step in scene understanding can be prohibitively expensive (3rd issue in introduction) for large data lifetimes.

Other related methods have also been developed but often customized to specific applications. For instance, Stringa {\em et al.}~\cite{Stringa98} describe a system to recover abandoned objects, Lee {\em et al.}~\cite{lee05} create a user interface to retrieve basic events such as the presence of a person, Meesen {\em et al.}~\cite{Meessen06} present an object-based dissimilarity measure to recover objects based on low-level features, and Yang {\em et al.}~\cite{Yang09} store human-body meta tags in a SQL table to humanoid shapes based on their skin color, body size, height, etc.

\subsection{Our Contributions}
In this paper, we extract a full set of low-level features as we have no a priori knowledge of what query will be asked. This is different from search procedures relying on high level information such as tracking~\cite{Hoferlin13} and human shape detection~\cite{Thornton11}.
Unlike scene understanding techniques, we have no training step; this would be incompatible with the data lifetimes and magnitude of the video corpus. Instead, we develop an approach based on exploiting temporal orders on simple features, which allows us to find arbitrary queries quickly while maintaining low false alarm rates.  We demonstrate a substantial improvement over scene-understanding methods such as \cite{Xiang08, Kuettel10} on a number of datasets in Section \ref{sec:results}.

\vspace{-0.2cm}
\section{Overview}
\label{sec:overview}
  
\begin{figure}[htp]
    \begin{center}
      \includegraphics[width=.85\linewidth]{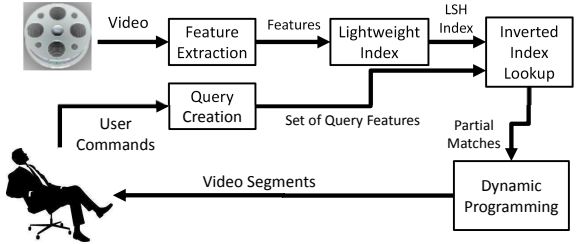}
    \end{center}
    \caption{\small From streaming video, low level features for each document are
      computed and inserted into a fuzzy, lightweight index.  A user inputs a
      query, and partial matches (features which are close to parts of the
      query) are inserted into a dynamic programming (DP) algorithm.  The
      algorithm extracts the set of video segments which best matches the
      query.\vspace{-0.2cm}}
    \label{fig:overview}
  \end{figure}

  To address the challenges of exploratory search, we utilize a two-step procedure 
  that first reduces the problem to the relevant data, and then reasons
  intelligently over that data.  This process is shown in Fig.
  \ref{fig:overview}.  As data streams in, video is pre-processed to extract
  low-level features - activity, object size, color, persistence and motion. 
  For CCTV footage, motion is estimated using Horn and Shunck approach ~\cite{Horn81}. 
  For Airborne footage, motion is estimated using tracklets. 
  These low-level features are hashed into a fuzzy, light-weight lookup table by
  means of LSH~\cite{Gionis99}.  LSH accounts for spatial variability and
  reduces the search space for a user query to the set of relevant {\em partial
  (local) matches}. 

  The second step is a search engine optimization which reasons over the partial
  matches to produce \emph{full matches}; segments of video which fit the entire
  query pattern, as opposed to part of it.  This optimization operates from the
  advantageous standpoint of having only to reason over the partial matches,
  which are the relevant subset of the video.  In surveillance video, where a
  long time can pass without relevant action, this dramatically reduces the
  workload of the optimization algorithm.
  
  We then present two solutions for finding full matches. The first is a greedy
  approach which flattens the query in time. The second, a novel dynamic
  programming (DP) approach, exploits the causal ordering of component actions
  that makeup a query. DP reasons over the set of partial matches and finds the
  best full match. We present two DP versions, one that uses optical-flow (for CCTV) and another that uses tracklets (for Airborne). We also show that as actions become more complex, the number of false positives in a surveillance
  environment goes to zero.
  
%
%
\vspace{-0.2cm}
\section{Feature extraction}
\label{sec:feature-extraction}
\vspace{-0.05cm}

In this section we describe the process of low-level feature extraction. We first explain the data structure used in our technique, then discuss feature extraction for CCTV and Airborne footage.   

\vspace{-0.2cm}
  \subsection{Structure}
  \label{sec:structure}

\begin{figure}[htb]
    \begin{center}
      \includegraphics[width=0.7\linewidth]{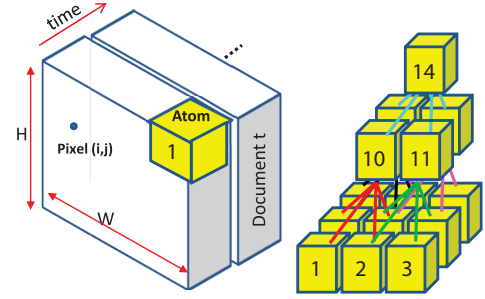}
    \end{center}
    \caption{\small (Left) Given an $W \times H \times F$ video, {\em documents}
      are non-overlapping video clips each containing $A$ frames.  Each of the
      frames are divided into {\em tiles} of size $B\times B$.  Tiles form an
      {\em atom} when aggregated together over $A$ frames. (Right) Atoms are grouped into trees - every adjacent set of four atoms is aggregated into a parent, forming a set of partially overlapping trees.\vspace{-0.3cm}}
    \label{fig:video-layout}
  \end{figure}

  For the purpose of feature extraction, a video is considered to be a
  spatio-temporal volume of size $H \times W \times F$ where $H \times W$ is the
  image size in pixels and $F$ the total number of frames in the video.  The
  video is divided along the temporal axis into contiguous {\em documents} each
  containing $A$ frames.  As shown in Fig. ~\ref{fig:video-layout}, each frame
  is divided into tiles of size $B \times B$.  An {\em atom} is formed by
  grouping the same tile over $A$ frames.  These values vary depending on the size of the video - for our videos, we chose B equal to 8 or 16 pixels, and A equal to 15 or 30 frames, depending on frame rate.  As the video streams in, features are
  extracted from each frame.  Whenever a document is created, each atom $n$ is
  assigned a set of features (see Sec.~\ref{sec:features} and Sec.~\ref{sec:features2}) describing the
  dynamic content over that region.

  We construct a pyramidal structure to robustify our algorithm to location and
  size variability in detected features. Each element of the pyramid has four
  children.  Since trees overlap, a $k$-level trees contains $M=\sum_{l=1}^k l^2$ nodes as shown in Fig.~\ref{fig:video-layout}.  In this approach, a
  document containing $U \times V$ atoms will be assigned $(U-k+1) \times (V-k+1)$
  partially overlapping trees to be indexed.  For instance, in Figure \ref{fig:video-layout}, we draw a depth-3 tree which aggregates $3 \times 3$ atoms.  This tree contains $M=14$ nodes.

  Each node of a tree is assigned a feature vector obtained by aggregating the
  feature vector of its  children nodes.  Let $n$ be a non-leaf node and $a, b, c, d$
  its four children.  The aggregation process can be formalized as
  \begin{equation*}
    \vec{x}_f^{(n)} = \psi_f
      \left(
        \vec{x}_f^{(a)}, \vec{x}_f^{(b)}, \vec{x}_f^{(c)}, \vec{x}_f^{(d)}
      \right),
  \end{equation*}

	where $\psi_f$ is an aggregation operator for the examined features. The resulting quantity is $\vec{x}_f^{(n)}$, a concatenation of all features.  
	Sec.~\ref{sec:features} and Sec.~\ref{sec:features2} present more details on the aggregation
  operator. Given that several features are extracted for each atom, aggregating a group
  of $k \times k$ atoms results in a set of feature trees $\set{\tree_f}$, one
  for each feature $f$. Given that a $k$-level tree contains $M$ nodes, each
  $\tree_f$ contains a list of $M$ feature instances, namely $\tree_f =
  \set{\vec{x}_f^{(n)}}$.
  
  \vspace{-0.2cm}
  \subsection{Feature Extraction for CCTV footage}
  \label{sec:features}

  As reported in the literature ~\cite{Meessen06,Stringa98,Yang09}, atom
  features can be of any kind such as color, object shape, object motion,
  tracks, etc.  We chose to use local processing due to the computational
  efficiency which makes it better suited to the constant data renewal
  constraint and real-time feature extraction.  Because our focus is surveillance
video, we assume a stable camera. 
Feature extraction computes a
  single value for each atom, which is aggregated into feature trees. Our method
  uses the following five features:


\noindent
  {\bf (1) Activity $x_a$:} Activity is detected using a basic background subtraction method~
  \cite{Benezeth10}.  The initial background is estimated using a median of the
  first 500 frames.  Then, the background is updated using the running average
  method.  At the leaf level, $x_a$ contains the proportion of active pixels
  within the atom.  Aggregation for non-leaf nodes in feature trees, $\psi_a$,
  is the mean of the four children.

\noindent
  {\bf (2) Blob Size $x_s$:} Moving blobs are detected using connected components analysis of the binary
  activity mask obtained from background subtraction \cite{Benezeth10}.  Blob size is the total
  number of active pixels covered by the connected component.  The aggregation
  operator $\psi_s$ for non-leaf nodes in feature trees is the median of
  non-zero children.  Whenever all four children have a zero size, the
  aggregation operator returns zero.

  \noindent
  {\bf (3) Color $\vec{x}_c$:} Color is obtained by computing the quantized histogram over every active pixel
  in the atom.  RGB pixels are then converted to the HSL color space.  Hue,
  saturation and luminance are quantized into 8, 4 and 4 bins respectively.  The
  aggregation operator $\psi_a$ for non-leaf nodes in feature trees is the
  bin-wise sum of histograms.  In order to keep relative track of proportions
  during aggregation, histograms are not normalized at this stage.
  
  \noindent
  {\bf (4) Persistence $x_p$:} Persistence is a detector for newly static objects.  It is computed by
  accumulating the binary activity mask obtained from background subtraction
  over time.  Objects that become idle for a long periods of time thus get a
  large persistence measure.  The aggregation operator $\psi_p$ for non-leaf
  nodes in feature trees is the maximum of the four children.
  
  \noindent
  {\bf (5) Motion $\vec{x}_m$:} Motion vectors are extracted using Horn and Schunck optical flow method
  ~\cite{Horn81}.  Motion is quantized into 8 directions and an extra ``idle''
  bin is used for flow vectors with low magnitude.  $\vec{x}_m$ thus contains a
  9-bin motion histogram.  The aggregation operator $\psi_m$ for non-leaf nodes
  in feature trees is a bin-wise sum of histograms.  In order to keep relative
  track of proportions during aggregation, histograms are not normalized at this
  stage.

  As mentioned previously, these motion features are extracted while the video
  streams in.  Whenever a document is created, its atoms are assigned $5$
  descriptors, namely $\set{x_a, x_s, \vec{x}_c, x_p, \vec{x}_m}$.  These
  descriptors are then assembled to form the $5$ feature trees $\set{\tree_a}$,
  $\set{\tree_s}$, $\set{\tree_c}$, $\set{\tree_p}$, $\set{\tree_m}$.  These feature
  trees are the basis for the indexing scheme presented in section
  ~\ref{sec:indexing}.  After feature trees are indexed, all extracted feature
  content is discarded, ensuring a lightweight representation.
  	It is worth noting that these features are intentionally simple.
This speeds up feature extraction and indexing while being robust
to small distortions due to the coarse nature of the features.  While motion can be sensitive to poorly-compensated camera motion or zoom, and color can be sensitive to illumination changes, the other features have been shown relatively robust to these effects \cite{Benezeth10}. In addition, we leverage on the dynamic programming in section \ref{sec:dynamic-programming} to limit false activity detection.

\vspace{-0.2cm}
\subsection{Feature Extraction for Airborne Footage}
  \label{sec:features2}
 
  In order to be robust to issues of low contrast and low frame-rate in airborne data, we extract tracklets and then derive features from them. Our tracklet extraction process uses a Viterbi-style algorithm with ideas from Pitie et al. \cite{pitie05} and Baugh et al. \cite{baugh09}.  At each frame, we generate a set of candidates to be tracked and update existing tracklets with these candidates based on distance in feature and position space.
 
\noindent
  
\begin{figure*}
	\centering
    \subfigure[]
		{
      \includegraphics[width=3.3cm,height=3.3cm]{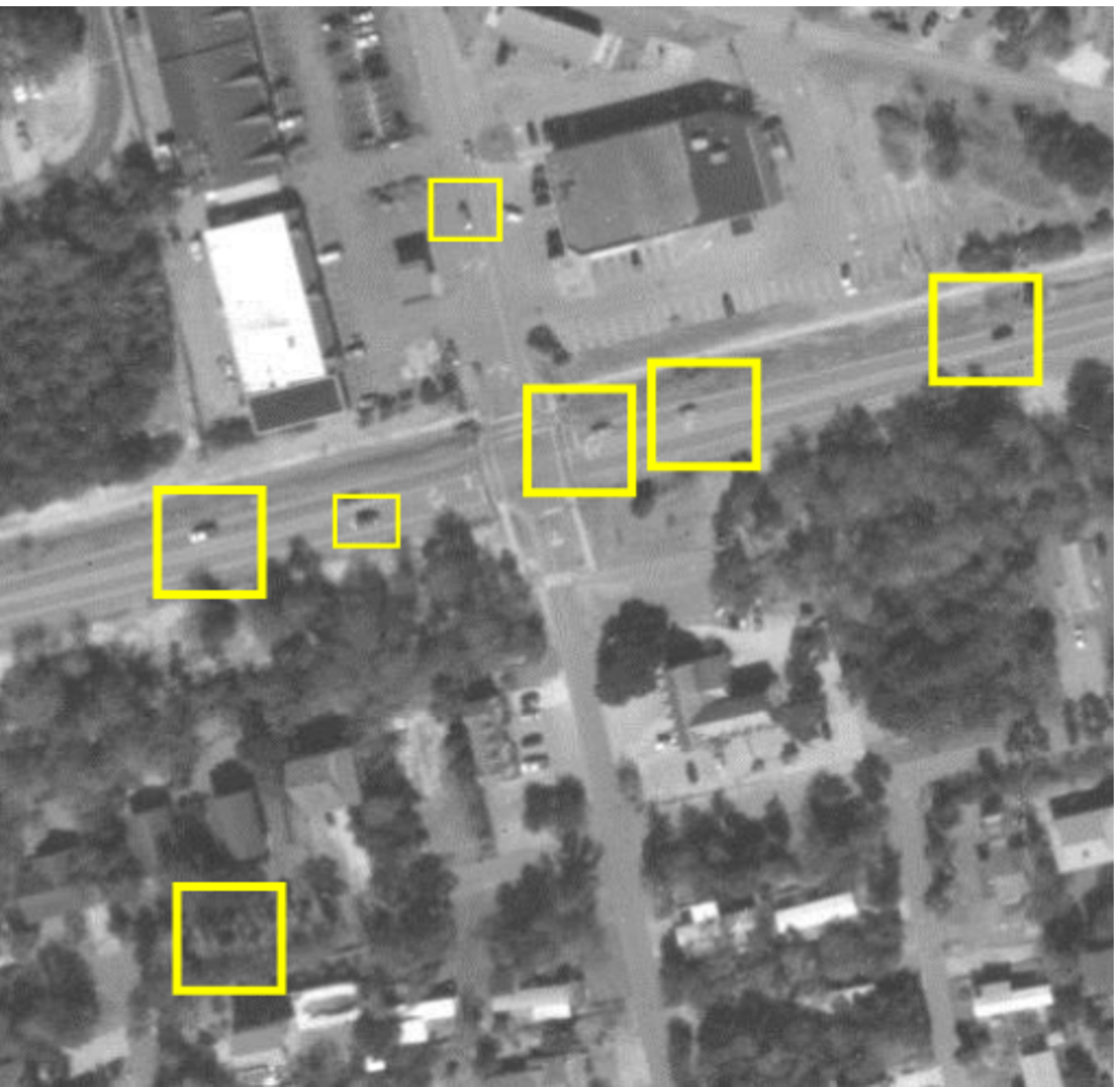}
    }
		\subfigure[]
		{
      \includegraphics[width=3.3cm,height=3.3cm]{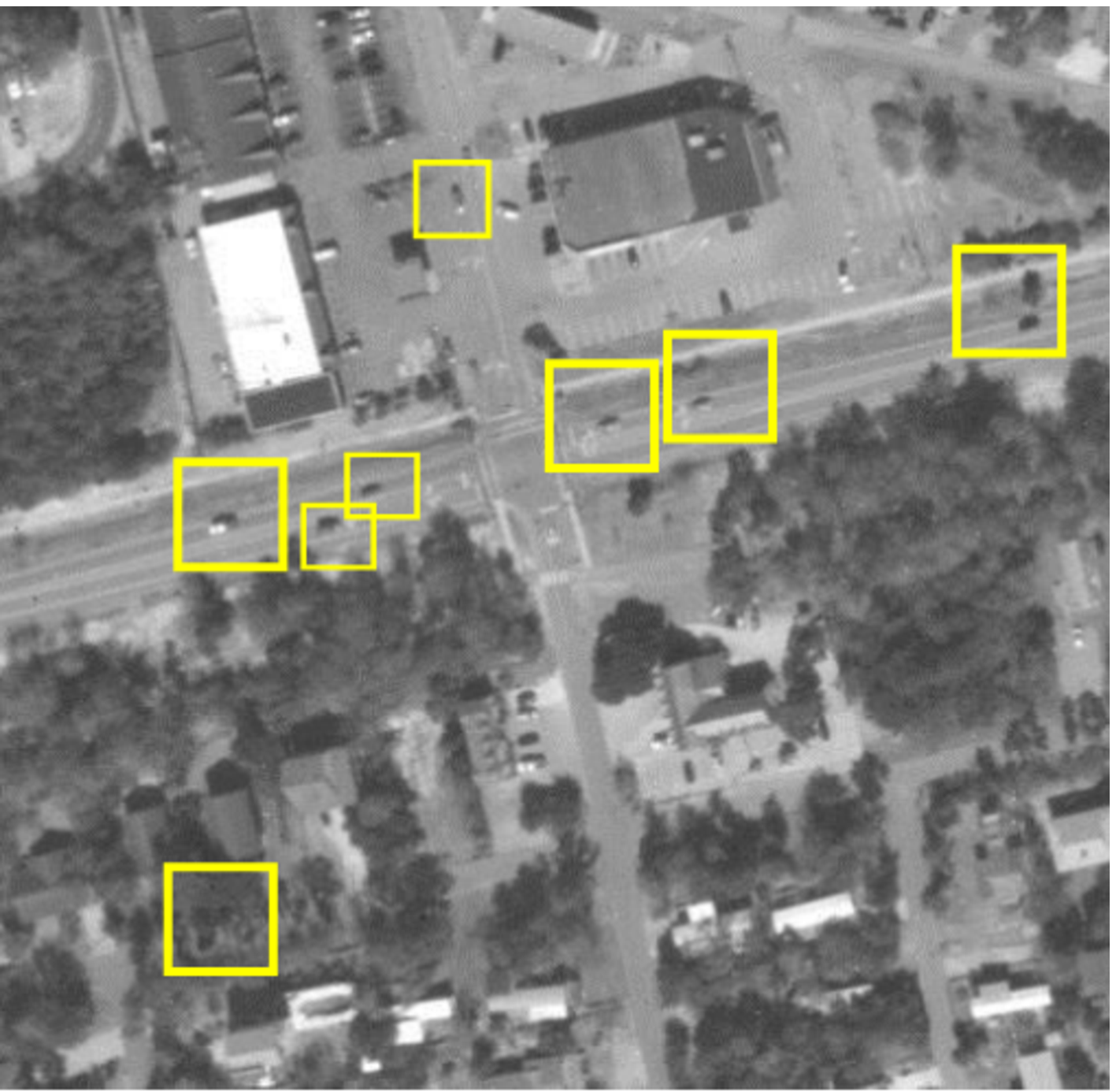}
    }
		\subfigure[]
		{
      \includegraphics[width=3.3cm,height=3.3cm]{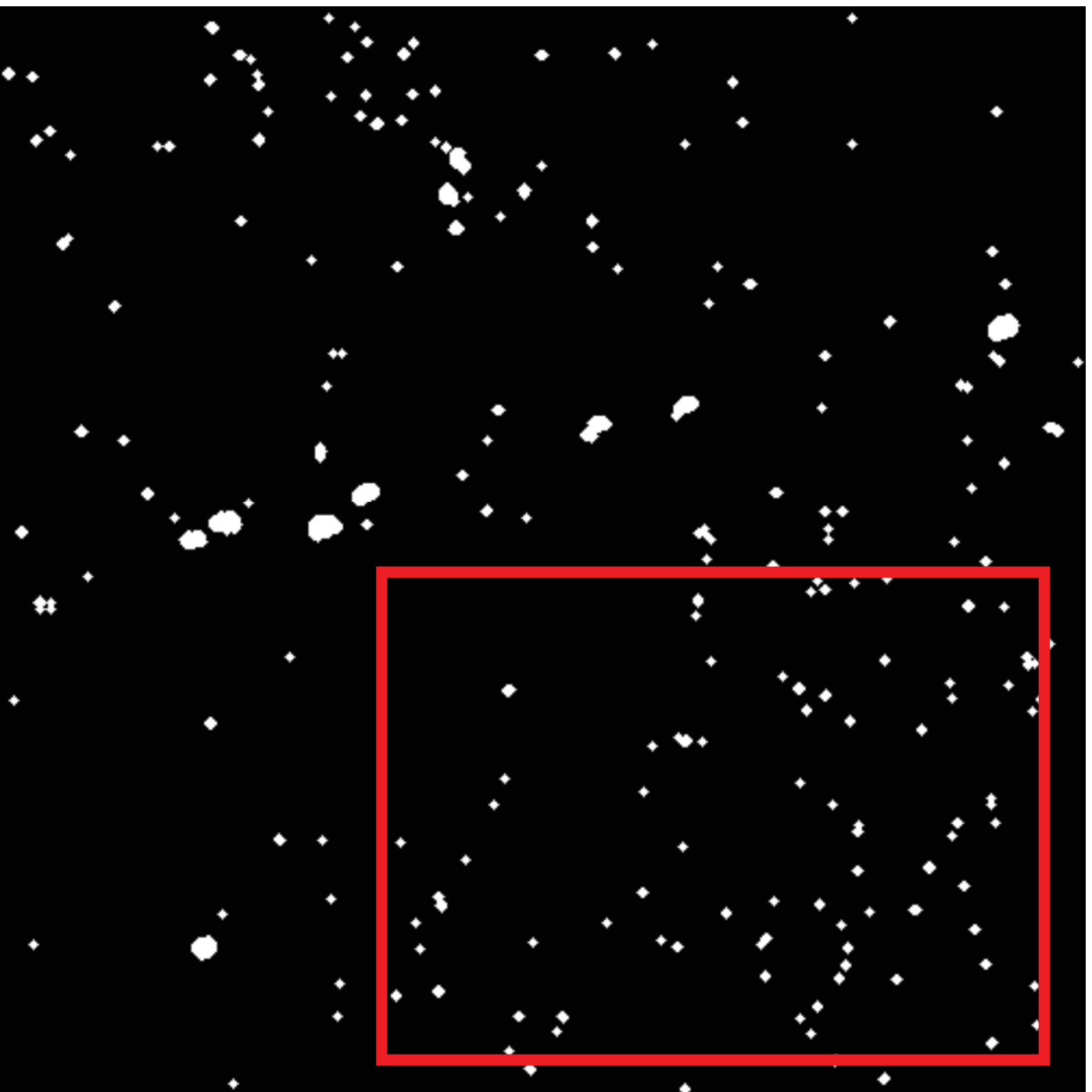}
    }
		\subfigure[]
		{
      \includegraphics[width=3.3cm,height=3.3cm]{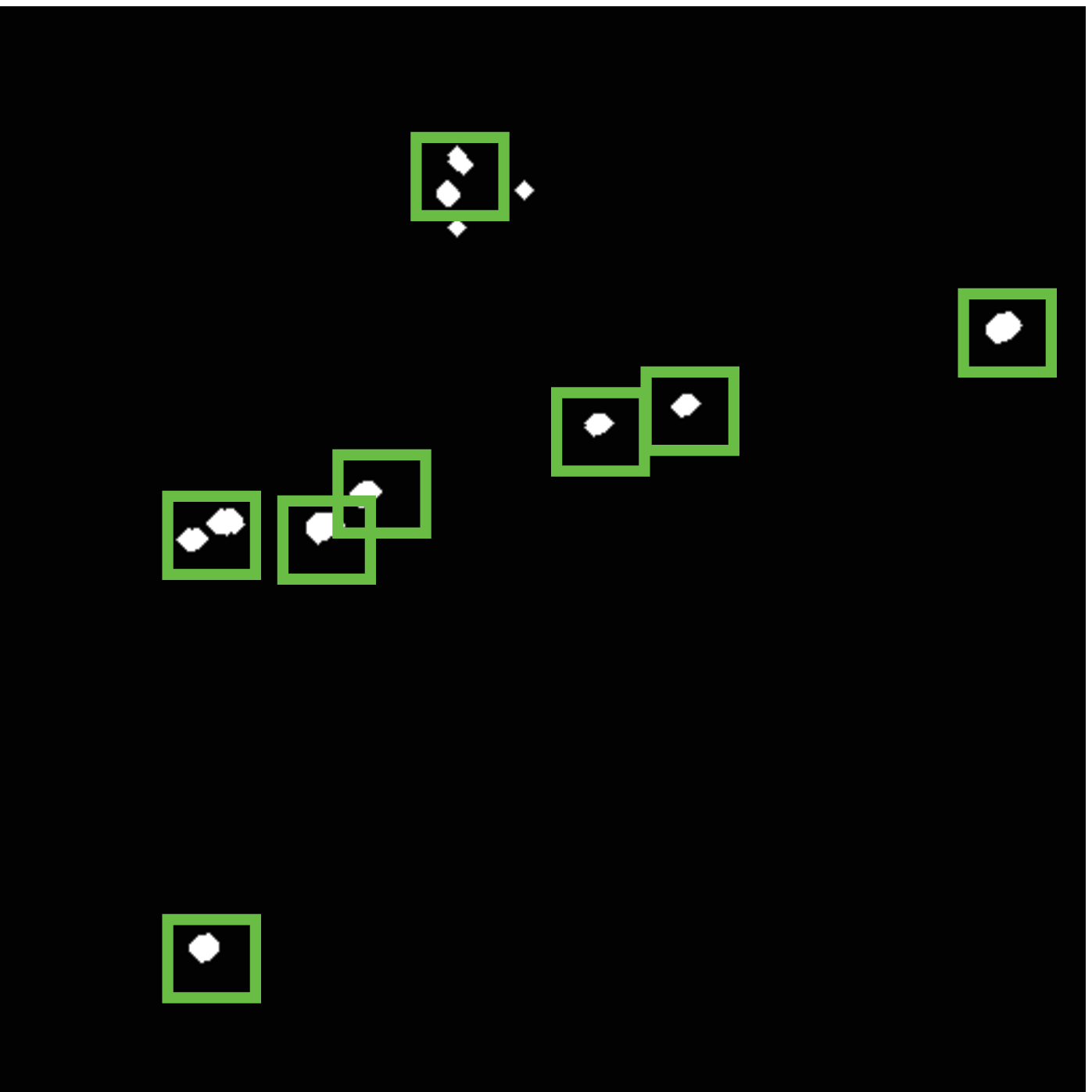}
    }\\
		\subfigure[]
		{
      \includegraphics[width=3.3cm,height=3.3cm]{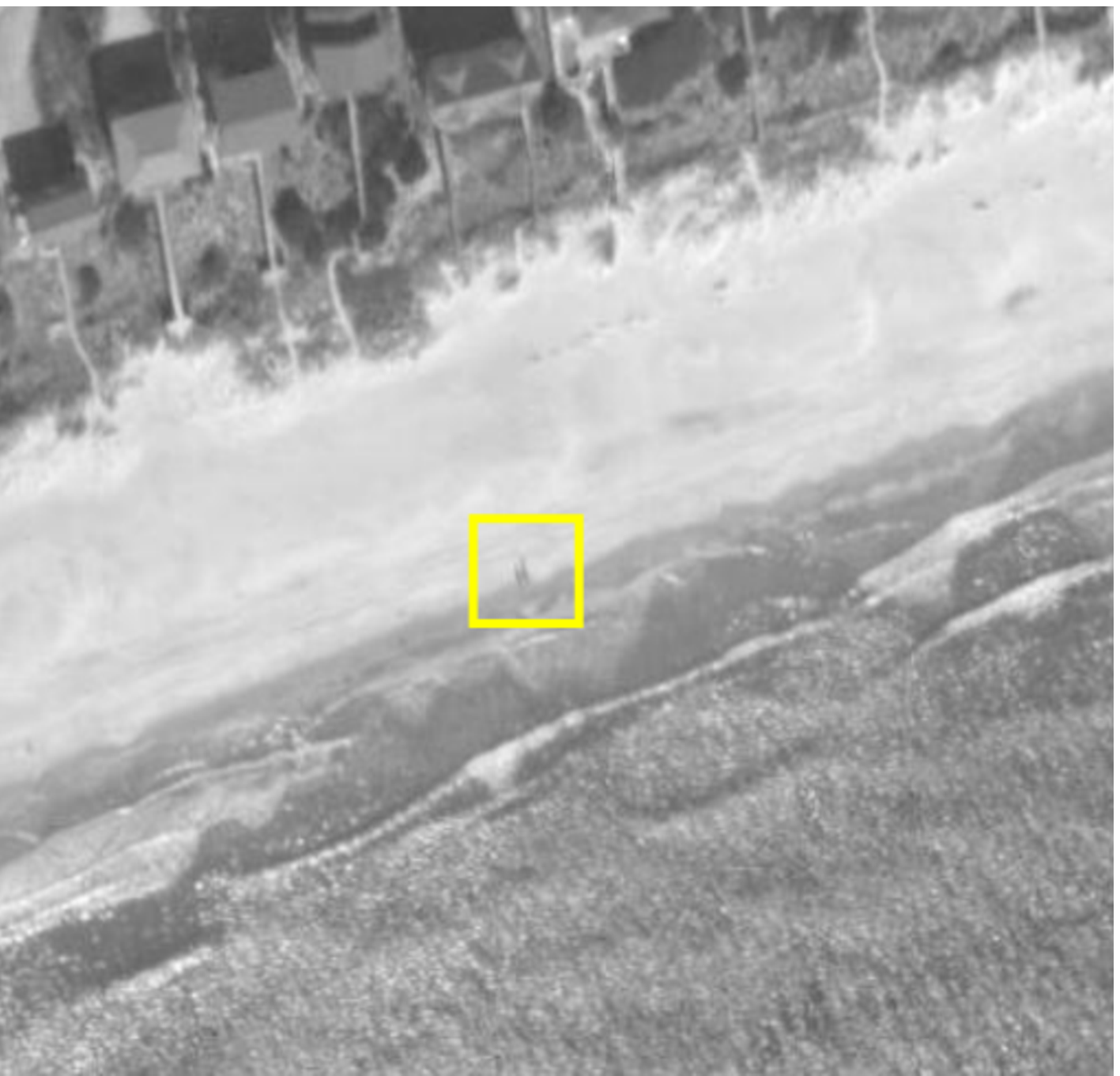}
    }
		\subfigure[]
		{
      \includegraphics[width=3.3cm,height=3.3cm]{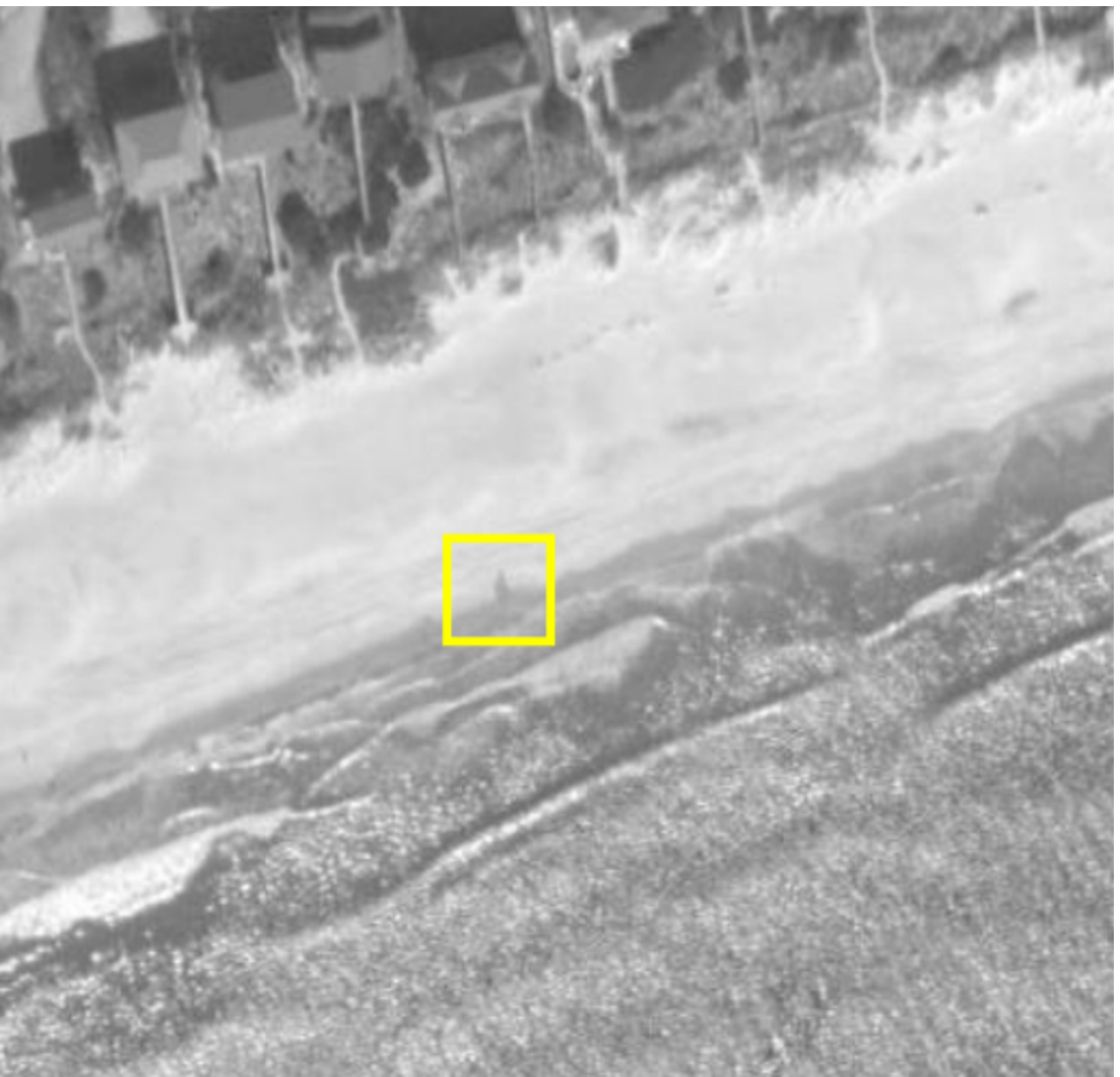}
    }
		\subfigure[]
		{
      \includegraphics[width=3.3cm,height=3.3cm]{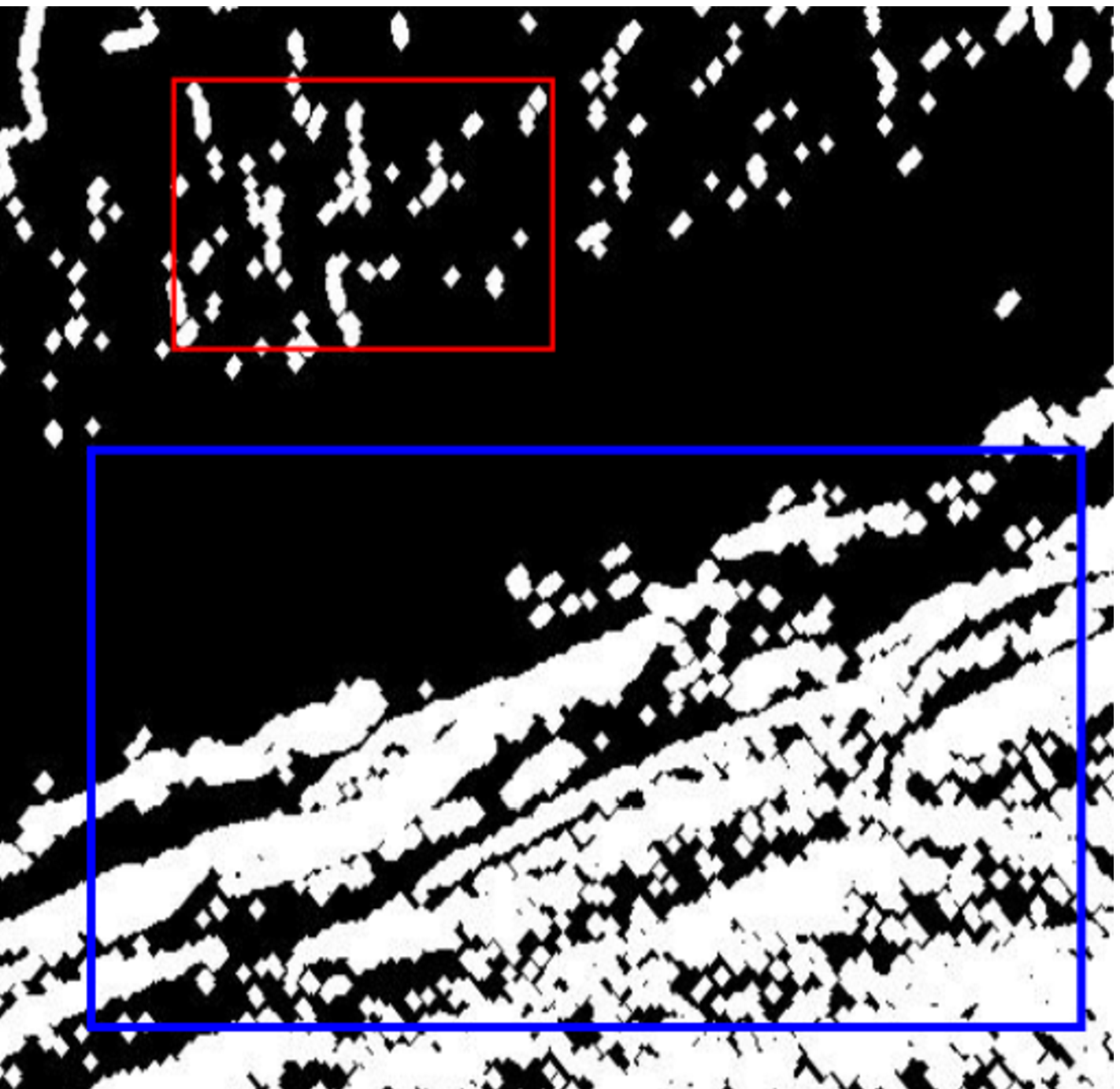}
    }
		\subfigure[]
		{
      \includegraphics[width=3.3cm,height=3.3cm]{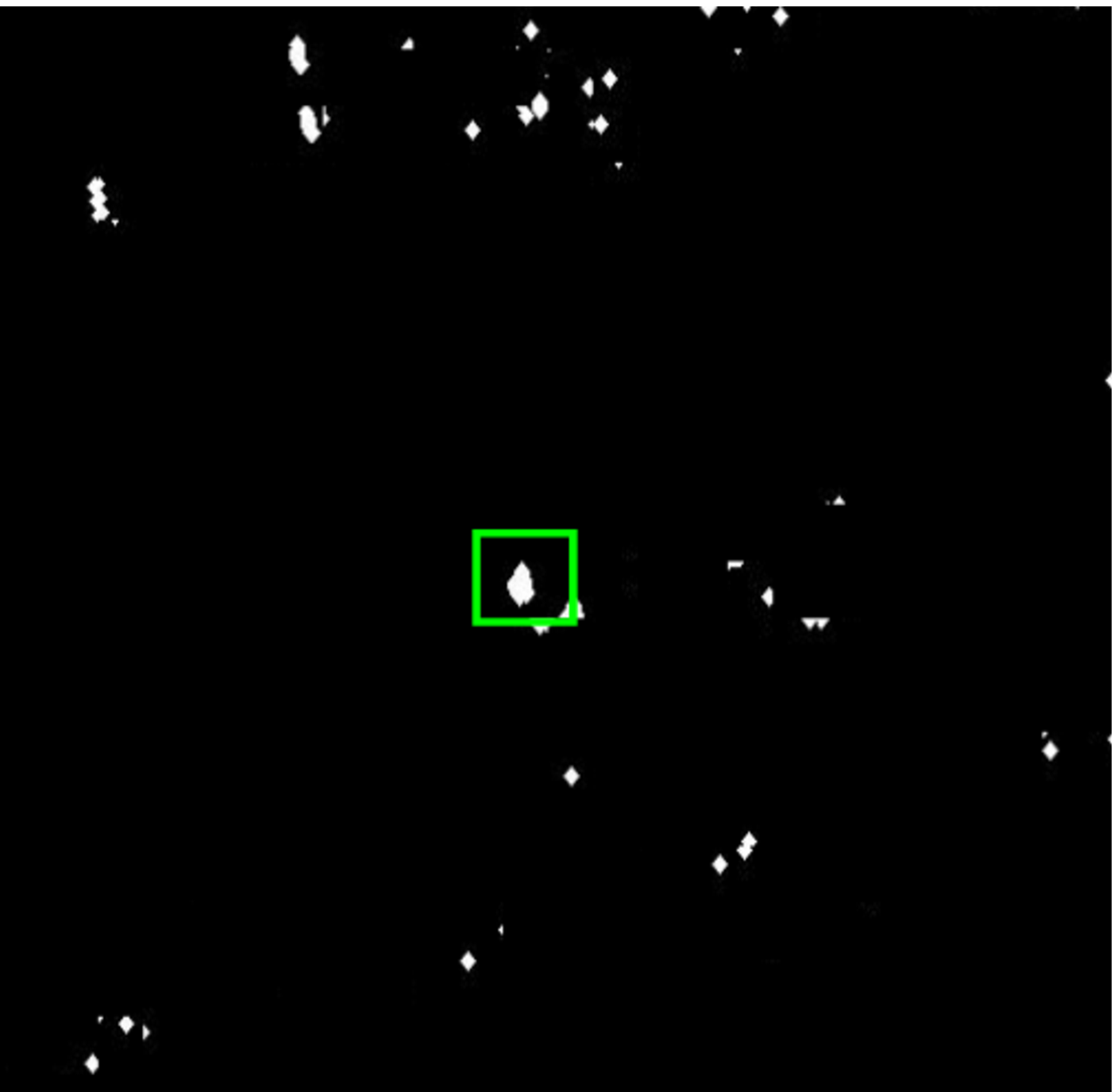}
    }
    \caption{\small First row, from left: Two consecutive frames showing cars (shown in yellow) moving near a junction, (c) Frame subtraction and (d) result after applying morphological opening. 
The final result (d) is void of noisy data shown in red in (c). Second row, from left: Two frames showing a group of people (shown in yellow) walking near a beach, (g) detection after applying morphological opening and (h) detected candidates of size more than 150 pixels in green.\vspace{-0.2cm} }
    \label{fig:BackSub}
  \end{figure*}

We are supplied with Airborne footage stabilized with respect to the global camera motion. In order to identify moving objects (cars and people) we use frame differencing. Consecutive frames are used for cars, and frames spaced by 10 for slower-moving humans. We use a very small detection threshold to ensure all false negatives are eliminated. Cars are detected if a frame difference of more than $15$ gray-scale levels is observed. A smaller threshold of $10$ is used for humans detection as they have lower contrast. Some detection errors often occur around borders of trees and objects (see Fig. \ref{fig:BackSub}, red region). Other artifacts are caused due to local motions as the ones generated by the moving sea waves (see Fig.\ref{fig:BackSub}, blue region). Such artifacts are removed through filtering by size. Here we first detect connected bodies and remove any body that contains more than $150$ pixels. 

An example of candidate selection with artifacts removal is shown in Fig.\ref{fig:BackSub}. Note that filtering by size eliminates the vast majority of false alarms. This is evident by examining the blue and red rectangles of Fig.~\ref{fig:BackSub} before and after filtering (third and fourth columns respectively). Quantitative results show that this filtering reduces false alarms by $72.3\%$ and $96.5\%$ for cars and humans respectively. This is taken as the average over processing $1000$ frames for each of Task 12 and 13 ( see table~\ref{tab:task1} ). It is worth noting that our algorithm only detects moving vehicles and humans. Stopped vehicles are dealt with by our dynamic programming algorithm which we introduce in Section~\ref{sec:dynamic-programming}. 


  Given a set of detection candidates, our goal is to associate them over time for the purposes of computing tracklets. To do so, we set up and solve a Viterbi Trellis as described in \cite{baugh09}. Given a set of detection candidates in frame $n$ and $n+1$, we estimate the temporal matching cost of each detection pair as follows	
  \begin{align}
  & \mathcal{C}_{l,m} = \Lambda_1 \| p_l - p_m \| + \Lambda_2 | z_l - z_m |  \nonumber
  \\ & + \Lambda_3 f_3(s_l,s_m) + \Lambda_4 f_4(c_l,c_m) + \Lambda_5 f_5(p_l,p_m).
  \label{eq:CostFun}
\end{align}

Here $l$ and $m$ denote the examined candidate of frame $n$ and $n+1$ respectively. $(p \; s \; z \; c)$ are the \underline{p}osition, \underline{s}hape, si\underline{z}e and \underline{c}olor of each examined candidate. $\Lambda_1 ... \Lambda_5$ are tunable weights to emphasize the importance of each term. Position is defined as the geo-registered location, shape is the binary mask of the occupied pixels and size is the number of pixels in the examined mask. The purpose of the five elements of Eq.~(\ref{eq:CostFun}) is to ensure that the target: (1) has not moved too far between $n$ and $n+1$, (2) is roughly the same size, (3) the same shape, (4) the same color and (5) it is roughly where we expect it to be given its previous motion and location. 

$s_l$ and $s_m$ are binary detection masks for the targets under examination, where $f_3(s_l,s_m)$ is the mean absolute error between both masks. Given the color measurements of the examined candidates $c_l$ and $c_m$, we fit a GMM for $c_l$ being $G(c_l)$ \cite{Bouman98}. We then estimate $f_4(c_l,c_m)$ as the error of generating $c_m$ from $G(c_l)$ \cite{Bouman98}. To calculate $f_5(p_l,p_m)$ we first estimate the expected location of $l$ in the next frame ($n+1$) given its current location in frame $n$. The expected location is estimated using a 3 frame window and a straight line, constant acceleration model. $f_5(p_l,p_m)$ is then taken as the absolute difference between the expected location of $l$ in $n+1$ and the actual location of $m$.

Given the matching costs for all possible $l$ and $m$ pairs, we approximate the two-dimensional assignment problem by greedy assignment for computational efficiency. That is we match the pairs with least matching cost first, remove them, match the next pair, remove them and keep doing so until all candidates in frame $n$ are matched with candidates in frame $n+1$. Note that this is a one-one assignment process hence no candidate in either frame $n$ or $n+1$ can have more than one match. This assignment, plus efficient computation of $f_4$, yield a simple tracking approach that can scale to large datasets.

\begin{figure}
    \begin{center}
      \includegraphics[width=.99\linewidth]{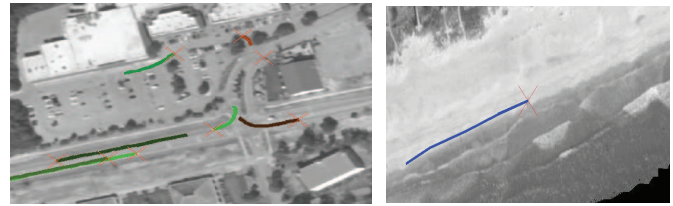}
    \end{center}
    \caption{\small Examples of tracklets generated by the technique discussed in Sec.~\ref{sec:features2}. Here tracks start with a red cross and their tails have different color. The example of the left shows tracklets generated for cars while the example on the right shows tracklet generated for a group of people walking on the beach.\vspace{-0.3cm}}     
    \label{fig:TracksExample}
  \end{figure}

Fig.~\ref{fig:TracksExample} shows examples of tracklets generated by the technique discussed in this section. Here we show tracklets of cars (see left of Fig.~\ref{fig:TracksExample}) and for a group of people walking on the beach (see right of Fig.~\ref{fig:TracksExample}). Note that our tracklets generator assumes temporal consistency and hence it disregards false detections that are temporally inconsistent. This further removes detection artifacts. In closing, note that we do not need particularly good long-term tracks; we need tracks good enough for our retrieval problem. 

%


\vspace{-0.2cm}
\section{Indexing \& Hashing}
\label{sec:indexing}

   When a document is created, features are aggregated into $(U-k+1) \times (V-k+1)$
  $k$-level trees.  For CCTV data, each tree is made of $5$ feature trees, namely
  $\set{\tree_a}, \set{\tree_s}, \set{\tree_c}, \set{\tree_p}, \set{\tree_m}$. For Airborne data there is one feature tree, namely $\set{\tree_d}$. $\set{\tree_d}$ is a 1-level tree that stores the current-to-next frame displacements for all points of the generated tracks. 
  
  To index a given feature tree $\tree_f$ efficiently, our method uses an inverted index for
  content retrieval.  Inverted index schemes, which map content to a location in
  a database, are popular in content-based retrieval because they allow
  extremely fast lookup in very large document databases.  For video, the goal
  is to store the document number $t$ and the spatial position $(u,v)$ in the
  database based on the content of $\tree_f$. This is done with a mapping
  function which converts ``$\tree_f$'' to an entry in the database where
  $(t,u,v)$ is stored. Two trees with similar contents, therefore, should be mapped to proximate locations in the index; by retrieving all entries which are near a query tree, we can recover the locations in the video of all features trees that are similar. Note that for Airborne data, in addition to storing $(t,u,v)$ for each tree, we also store the track ID. This encourages results to be generated from as few tracks as possible and hence makes solution more robust to clutter.

  This mapping and retrieval can be made for which update and lookup exhibit flat performance ($O(1)$ complexity).  Because similar content at different times is mapped to the same bin, the time required to find the indices of matching trees does not scale with the length of the video.  Obviously, the total retrieval time must scale linearly with the number of matching trees, but this means that the run-time of the retrieval process scales only with the amount of data which is relevant.  In videos where the query represents an infrequently-performed action, this optimization yields an immense improvement in runtime.


    \noindent
  {\bf Hashing:} A hash-based inverted index uses a function to map content to an entry in the
  index.  This is done with a hashing function $h$ such that $h: \tree_f \to j$,
  where $j$ is a hash table bucket number.  
%
  In this paper, we use a locally-sensitive hashing (LSH) function~\cite{Gionis99}.  LSH is a technique for approximation of nearest-neighbor search.
  In contrast to most hashing techniques, LSH attempts to cluster similar
  vectors $\vec x$ and $\vec y$ by maximizing the probability of collisions for descriptors within a
  certain distance of each other:
  \begin{equation*}
    \vec{x} \approx \vec{y} \implies \prob{h(\vec{x})=h(\vec{y})} \gg 0.
  \end{equation*}
  If feature trees are close in Euclidean distance (the element-wise square of the distances between node values in the two trees is small), then the probability of them
  having the same hash code is high.  Because our feature trees contain $M$
  real-valued variables, LSH functions can be drawn from the p-stable family:
  \begin{equation*}
    h_{\vec{a},b,r}(\tree_f)
      = \floor{\frac{\vec{a}\cdot\tree_f+b}{r}},
  \end{equation*}
  where $\vec{a}$ is a $M$-dimensional vector with random components drawn from
  a stable distribution, $b$ is a random scalar drawn from a stable distribution
  and $r$ is an application-dependent parameter.  Intuitively, $\vec{a}$
  represents a random projection, an alignment offset $b$ and a radius $r$
  controlling the probability of collision inside the radius.

  Indices $I_f$ are built and searched independently for each feature.  Thus, the
  database is made of five indices for the CCTV data and two indices for the Airborne data. Here there is one database index for each feature $f$.  Each index
  $I_f$ is composed of a set of $n$ hash tables $\set{T_{f,i}},~\forall
  i=1,\hdots,n$.  Each hash table is associated to its own hash function $H_{f,i}$
  drawn from the p-stable family $h_{\vec{a},b,r}$.  The parameter $r$ can be
  adjusted to relax or sharpen matches.  In our implementation, $r$ is fixed
  for a given feature.  The random parameters $\vec{a}$ and $b$ ensure projections
  from the different hash functions complement each other.

  Given a feature tree $\tree_f$ at location $u,v$ with hash code
  $H_{f,i}(\tree_f,u,v)=j$, $T_{f,i}[j]$ denotes the set of document numbers
  $\set{t}$ such that feature trees at $(t,u,v)$ have similar content. Lookup in
  the index $I_f$ consists of taking the union of document numbers returned by
  lookup in all tables $\set{T_{f,i}}$:
  \begin{equation*}
    I(\tree_f,u,v) = \displaystyle\cup_{i=1}^{n}
      T_{f,i}\left[H_{f,i}(\tree_f),u,v\right].
  \end{equation*}

  \begin{figure}[htb]
    \begin{center}
    (a) \includegraphics[width=0.42\linewidth]{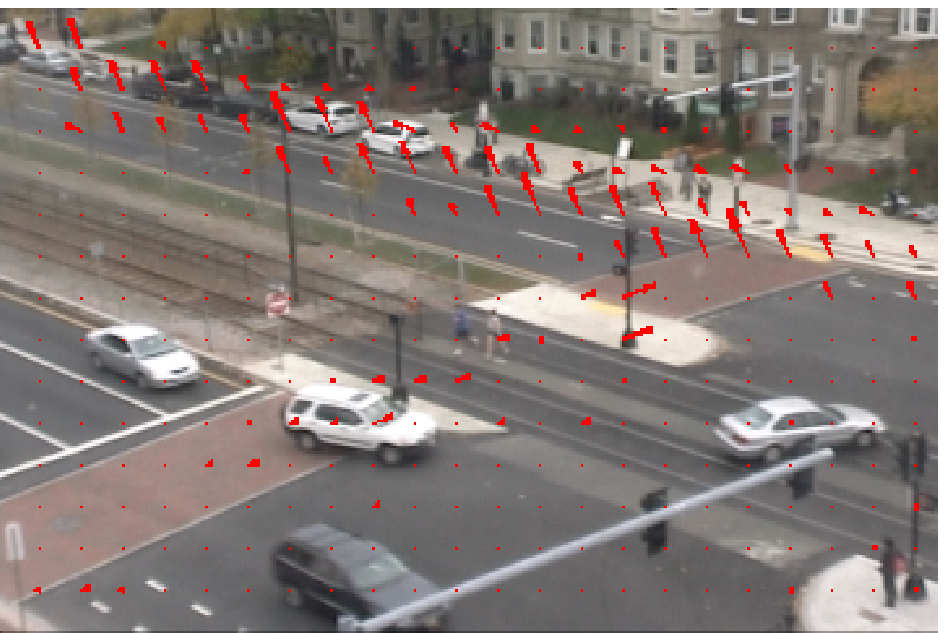}
    (b) \includegraphics[width=0.42\linewidth]{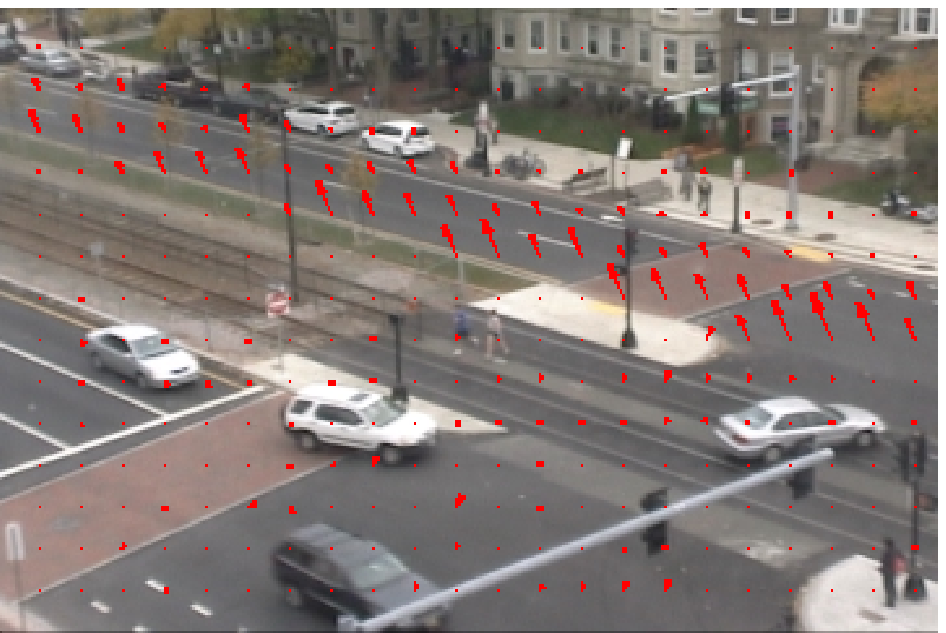}
    \\
    (c) \includegraphics[width=0.42\linewidth]{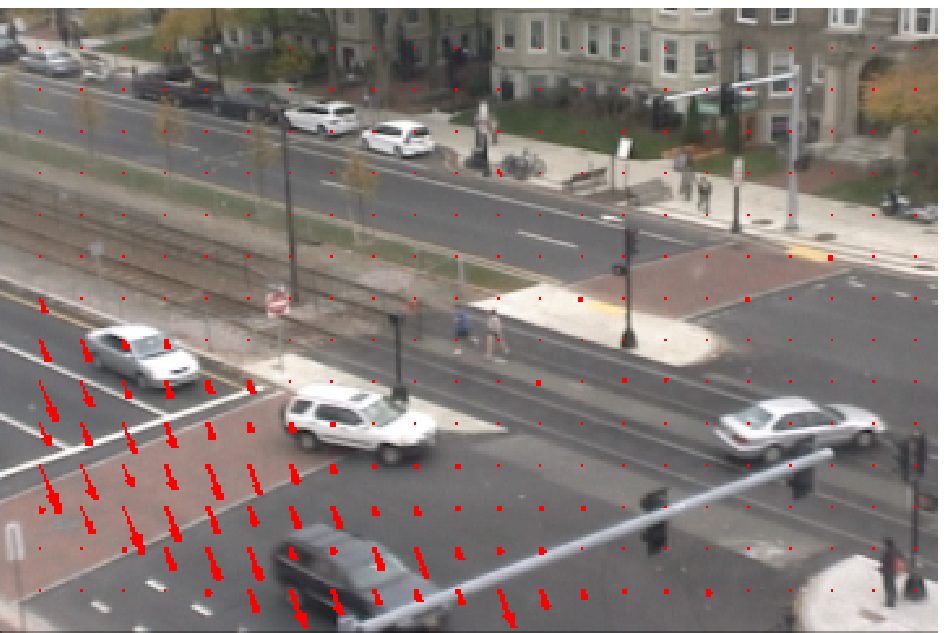}
    (d) \includegraphics[width=0.42\linewidth]{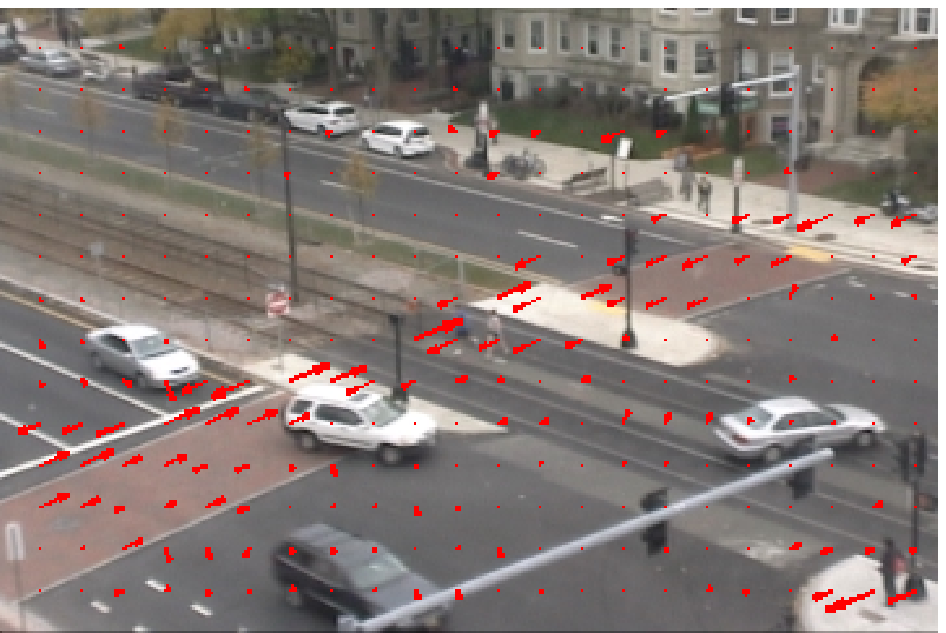}
        \end{center}
    \caption{\small Contents of four buckets of a hash table for the
      {\em motion} feature.  Arrows size is proportional to the number of hits
      at the specified location across the entire video.  The hash buckets are
      associated to activity (a) side walk (b) upper side of the
      street (c) lower side of the street (d) crosswalk.\vspace{-0.2cm}}
    \label{fig:collisions}
  \end{figure}

  Fig. ~\ref{fig:collisions} illustrates several feature trees partitioned
  into groups, where trees in the same group have been given the same hashing
  key.  For a given video, we plotted the content of four of the most occupied
  buckets for the motion feature $\tree_m$.  As one can see, the trees
  associated to similar motion patterns in various parts of the scene have been
  coherently hashed into similar buckets.

%
%
  %
  %

  {\bf Lightweight Storage:} In contrast to approaches relying on a distance
  metric, such as K-nearest neighbor search, the hash-based index representation
  does not require storage of feature descriptors.  $T_{f,i}$ contains only
  document numbers $\set{t}$, which are stored in $4$-byte variables.  Our features, being local and primitive, are dependent on foreground content.  As such, both our indexing times and storage scale linearly with the amount of foreground content in the video.  This is a useful feature for surveillance video, which can have persistently inactive spaces or times throughout a video.  For example, the size of the index for a 5 hour video with an activity rate of $2.5\%$ is only
  150kb while the input color video requires almost 7Gb (compressed).

  \noindent
  {\bf Building Lookup Table:} As video streams in, the index for each feature
  is updated by a simple and fast procedure.  After extraction of feature $f$
  for document $t$ is completed, the extracted features are grouped into trees,
  as described in Sec. \ref{sec:feature-extraction}.  Then, $I_f$ is updated
  with the mapping $\tree_f \to (t,u,v)$ for each tree position $(u,v)$ covering
  an area with significant activity.  This is repeated for each feature $f$.


\section{Search engine}
\label{sec:search-engine}

  We showed how to extract low-level features from a video sequence,
  bundle them into trees and index them for $O(1)$ content-based retrieval. Here
  we explain how to use feature index for high-level search.

\vspace{-0.3cm}
\subsection{Queries}
\label{sec:queries}

\begin{figure}[h!]
    \begin{center}
      \includegraphics[width=1 \linewidth]{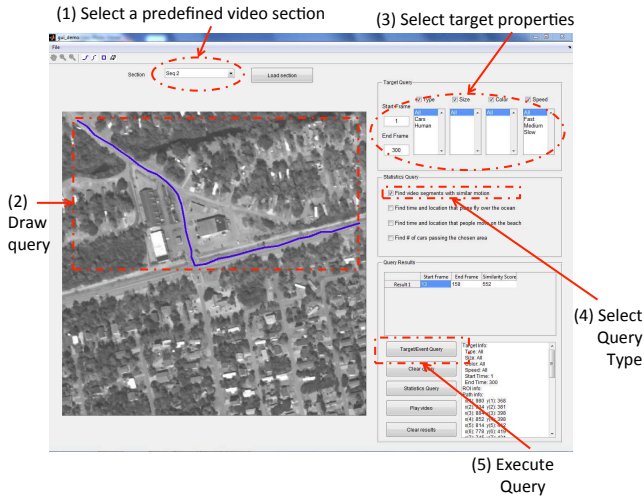}
    \end{center}
    \vspace{-.3in}
    \caption{\small The query creation GUI provides a straightforward way to construct queries.  The user draws each action component (shown in blue), and can additionally specify features.\vspace{-0.4cm}}
    \label{fig:GUI}
 \end{figure}

	In video search without exemplars, it is essential to provide a straightforward way for a user to input a query.  For our , a query is defined as an ordered set of \emph{action components} (for simple queries, a single component frequently suffices), each of which is a set of feature trees.  To create a query, the user types in the number of action components, and is then presented with a GUI, shown in Figure \ref{fig:GUI} containing the first frame of the video to search for each of the action components.  The user then selects the type of feature he wishes to search for, and draws the region of interest (ROI) that he wishes to find it in. These regions and the features (directions of motion, in this case) are shown in Fig.\ref{fig:results} as green areas and red arrows, respectively.
	
	As an example, to describe a u-turn, the user might describe three action components:  one detecting motion approaching an intersection, one detecting motion turning around, and one detecting motion leaving the intersection.  Likewise, a man hopping a subway might be represented by somebody approaching the subway, jumping up, and then continuing past the subway.
	
	Because our features (activity, size, color, persistence and motion) are semantically meaningful, this method of input is a relatively accessible way to define a query.  After the features and the ROI are selected, a feature tree $tree_f$ is created to represent that action component.	Note that this formulation provides the user with a way to produce a complex query vocabulary.  A single component could describe a search for ''small red stationary blob'' or 'large blob moving to the right''.  While our claims to simplicity are theoretical, they are also supported by anecdotal evidence. People unfamiliar with the system are able to create their own queries in under a minute after a brief explanation of the tools.
	
	After the action component has been input, the set of feature trees is extracted from it using the formulation in Section \ref{sec:feature-extraction}.  These feature trees is used to query the inverted index of Section \ref{sec:indexing} to produce a set of documents and locations called \emph{partial matches} $M(q)$.  $M(q)$ is a list of spatio-temporal positions $(u,v,t)$ which contain similar trees to query $q$.  In the case where the query contains multiple feature trees, the set of matching locations is the intersection of the matches for individual trees. Fig.\ref{fig:results} presents 10 queries with their ROI.

  \vspace{-0.2cm}
  \subsection{Full matches}
  \label{sec:full-matches}

  Search in a surveillance video requires more than partial matches.  Activities
  in a video are inherently complex and show significant variability in time
  duration.  For instance, a fast car taking a U-turn will span across fewer
  documents and generate different motion features than a slow moving car. Also,
  due to the limited size of a document (typically between $30$ to $100$
  frames), partial matches may only correspond to a portion of the requested
  event.  For example, partial matches in a document $t$ may only correspond to
  the beginning of a U-turn.  The results expected by a user are so-called {\em
    full matches}, i.e. video segments $[t,t+\Delta]$ containing one or more
  documents ($\Delta > 0$).  For example, the video segment $R =
  \set{t,t+1,t+2}$ corresponds to a full U-turn match when documents $t,t+1,t+2$
  contain the beginning, the middle and the end of the U-turn.  Given a query
  $q$ and a set of partial matches $M(q)$, a full match starting at time $\tau$
  is defined as
  \begin{eqnarray}
    \label{eq:value-function}
    R_{q,\tau}(\Delta)=\set{(u,v)|(t,u,v) \in M(q),~ \forall t \in
      [\tau,\tau+\Delta]}.
  \end{eqnarray}

  Thus, $R_{q,\tau}(\Delta)$ contains the set of distinct coordinates of trees
  partially matching $q$ in the video segment $[\tau,\tau+\Delta]$.

  We propose two algorithms to identify these full matches from the set of
  partial matches.  The first is a greedy optimization procedure based on the total number of partial matches in a document that does not exploit the temporal ordering of a query.  The second approach (Sec.~\ref{sec:dynamic-programming}), uses dynamic programming to exploit temporal structure of the  query action components.

  \subsubsection{Full matches using a greedy algorithm}
  \label{sec:greedy-algorithm}

  The main difficulty in identifying full matches comes with the inherent
  variability between the query and the target.  This includes time-scaling,
  false detections and other local spatio-temporal deformations.  Consequently,
  we are faced with the problem of finding which documents to fuse into a full
  match given a set of partial matches. We formulate it in terms of the
  following optimization problem:
  \begin{equation}
  \label{eq.opttau}
    \Delta^* = \arg\max_{\Delta > 0} v_{q,\tau}(\Delta)
  \end{equation}
  where $q$ is the query, $\tau$ is a starting point and $\Delta$ the length of
  the retrieved video segment.  The value function $v_{q,\tau}(\Delta)$ maps the
  set of partial matches in the interval $[\tau,\tau+\Delta]$ to some large
  number when the partial matches fit $q$ well and to a small value when they do
  not.  To determine the optimal length of a video segment starting at time
  $\tau$, we maximize the above expression over $\Delta$.

  While many value functions are viable, depending upon user preference, a
  simple and effective $v_{q,\tau}(\Delta)$ is:
  \begin{eqnarray}
    \label{eq:opt}
    v_{q,\tau}(\Delta) =& \exp(|R_{q,\tau}(\Delta)|-\lambda\Delta),
  \end{eqnarray}
  where $R_{q,\tau}(\Delta)$ is defined by Eq. \eqref{eq:value-function} and
  $|R_{q,\tau}(\Delta)|$ is the total number of distinct matching locations
  found in the interval $[\tau, \tau + \Delta]$.  The value function is
  time-discounted since $R_{q,\tau}(\Delta)$ is increasing in $\Delta$ (by
  definition, $R_{q,\tau}(\Delta) \subseteq R_{q,\tau}(\Delta+1)$).  The
  parameter $\lambda$ is a time-scale parameter and loosely controls size
  of retrieved video segment.

  We can determine $\Delta$ by a simple and fast greedy algorithm. The algorithm
  finds a set of non-overlapping video segments and a natural ranking based the
  value function provided above.  As will be shown in Sec.
  ~\ref{sec:results}, Eq. (\ref{eq:opt}) produces compact video segments while
  keeping low false positives and negatives rates.

  It may seem strange that such a simple value function provides accurate
  results over a wide range of queries and videos.  Intuitively, this is because
  the more complex a query is, the less likely it is to be generated by
  unassociated random actions.  Specifically, given a query $Q$ containing $N$ components drawn from dictionary $D$, the probability $P(Q)$ of a randomly drawing $Q$ as sequence of video components within $\Delta$ frames is given  by Eq. \eqref{eq:bounds}, using Stirling's approximation~\cite{Abramowitz2002}.
  
  \begin{equation}
  \label{eq:bounds}
  \displaystyle P(Q) =\binom{|D|}{N} \frac{|D|^{\Delta - N}}{|D|^{\Delta}} = \binom{|D|}{N} {|D|}^{-N} \approx e^{-N \log \frac{|D|}{\Delta}}
  \end{equation}
  
  The expression shows that if the dictionary is large relative to the time-interval $\Delta$ then the probability of randomly drawing the user defined query decreases exponentially with the number of query components. Nevertheless, we can significantly improve this bound if we also account for the order of the query components. 
  \subsubsection{Full matches with Dynamic Programming(DP)}
  \label{sec:dynamic-programming}

  \begin{figure}
    \begin{center}
      \includegraphics[width=0.99 \linewidth]{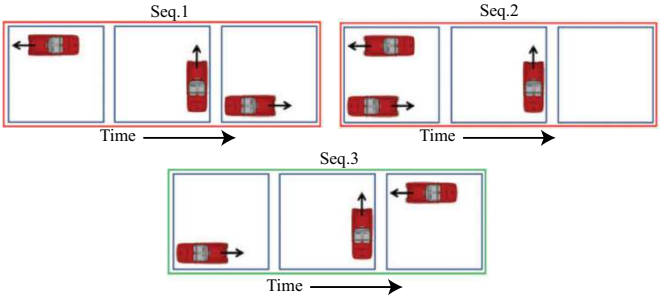}\vspace{-0.4cm}
    \end{center}
    \caption{\small Three sequences of actions. All have equivalent values of
      $R_{Q,\tau}(\Delta)$; only seq.3 contains a valid U-turn.\vspace{-0.3cm}}
    \label{fig:causality}
    \vspace*{-0.1in}
  \end{figure}

  We can improve upon the performance of the greedy algorithm in Sec.~\ref{sec:greedy-algorithm} by exploiting the order of the action components.  For example, in the
  example of a man hopping a subway turnstile, he has to approach the turnstile
  from the wrong direction, hop over it, and continue.  For a car taking a
  u-turn, it has to approach the intersection, turn across it, and depart the
  way it came.  While these component actions have many valid configurations when examined independently, there is only one order in which they represent the desired full action.  This is illustrated in Fig.~\ref{fig:causality}.

  In the greedy optimization of Section \ref{sec:greedy-algorithm}, we ignore the time-ordering that a set of queries contains.  The value function
  $R_{q,\tau}(\Delta)$ does not differentiate between the sequences of actions \emph{(Forward, Left, Back)} and the sequence of actions \emph{(Back, Forward, Left)}.  Intuitively, this should hurt performance - false matches are more likely to appear than were we to exploit this causality.  We quantify the improvement upon the performance bounds given in Eq. \eqref{eq:bounds} in Eq. \eqref{eq:dpbounds}.
  
  \begin{equation}
  \label{eq:dpbounds}
  \displaystyle P(Q)=\frac{e^{-N \log {\frac{|D|}{\Delta}}}}{N!}  \approx e^{ -N\log(\frac{|D|N}{\Delta})}
  \end{equation}

It is worth highlighting the difference between the two expressions Eq.~\ref{eq:bounds} and Eq.~\ref{eq:dpbounds}. In the latter expression we only need $|D|N$ to be sufficiently large relative to $\Delta$ as opposed to $|D|> \Delta$. Consequently, if we were to account for the order of query components, the probability of randomly matching the user defined query approaches zero even when $\Delta$ scales with the size of the dictionary.

	After a user created a query containing a set of $N$ action components (with GUI described in Section \ref{sec:queries}), we search through the inverted index to retrieve $N$ sets of matching locations within the video, one for each action component.

  After this pre-processing step, we adapt the Smith-Waterman dynamic programming algorithm
  \cite{SW81} , originally developed for gene sequencing, to determine the best set of partial matches in the query.  Our algorithm,
  described in Algorithm \ref{alg:dynamic-programming}, operates over the set of
  matches $m_{\tau,\alpha}\in M(q)$ which contains matches in document $\alpha$ to
  action component $\tau$ to recover a query that has been distorted.  For the
  purposes of our videos, we consider three types of distortion, namely
  insertion, deletion and continuation. 

\noindent
{\bf (1) Insertion} covers documents where we believe the query is happening but
  there are no partial matches.  This can happen if unrelated documents are
  inserted, if there is a pause in the execution of the activity, or if the
  activity is obscured. \\
{\bf (2) Deletion} covers documents where sections of the query are missing.  If a
  deletion has occurred, one (or more) of the action components in the query
  will not be present in the video.  Deletions can happen because of
  obscuration, or because somebody does not perform one component of the
  full action. \\
{\bf (3) Continuation} covers documents where we continue to see an action
  component which we have already seen.  This is important because of time
  distortion; a single action component does not necessarily occur in
  multiple consecutive documents before the next action component is reached.
	
    \begin{algorithm}[ht!]
      \caption{Dynamic programming (DP) algorithm}
      \label{alg:dynamic-programming}
      \begin{algorithmic}[1]
        \Procedure{Search}{m, W, T}
          \State $V \gets 0$; $paths \gets \emptyset$; $\tau \gets 1$; $\alpha \gets 1$
          \While{$\tau \le $ number of documents}
            \While{$\alpha \le $ number of action components} 
                           
              \State $V_{\tau,\alpha} \gets \max
                \left\{
                \begin{array}{c}
            0 \\
            (V_{\tau-1,\alpha-1} + W_{M})*m_{\tau, \alpha}        \\
            (V_{\tau-1,\alpha} + W_{C})*m_{\tau, \alpha}   \\
            (V_{\tau-1,\alpha} + W_{D})*(1-m_{\tau, \alpha})   \\
            (V_{\tau,\alpha-1} + W_{I})*(1-m_{\tau, \alpha})  \\
                \end{array}
                \right . $
                
             \If{Airborne Data}
               
               \State $V_{\tau,\alpha} \gets  V_{\tau,\alpha} + {\bf 1}_{\phi_{\tau-1, \alpha-1}}(\phi_{\tau, \alpha})*m_{\tau, \alpha}*$
							 \State $W_{S}$
                
             \EndIf
             
                \State Let $(a,b)$ be the index which was used to \State generate the maximum value
                \If{$V_{(\tau,\alpha)} > 0$}
                  \State $paths_{\tau,\alpha} \gets paths_{a,b} \cup (\tau,\alpha)$
                \Else
                  \State $paths_{\tau,\alpha} \gets paths_{a,b}$
                \EndIf
                \State $\alpha \gets \alpha + 1$
            \EndWhile
      \State $\tau \gets \tau + 1$
          \EndWhile

          \State $Matches \gets \emptyset$
          \While{$max(V) > T$}
          	\State Let $(a,b)$ be the index of $V$ containing the \State maximum value
          	\State $Matches \gets Matches \cup paths_{a,b}$
          	\For{$\tau, \alpha \in paths_{a,b}$}
          		\State $V_{\tau, \alpha} = 0$
          	\EndFor
          \EndWhile
          \State Return $Matches$, the set of paths above threshold T 
        \EndProcedure
      \end{algorithmic}
    \end{algorithm}
	
  In addition to the three above distortions, we introduce an extra term to handle the Airborne data. This term adds more score to a match if its Track ID is the same as the ID of the match in the previous document. In Algorithm \ref{alg:dynamic-programming}, the track ID of an examined cell is $\phi_{\tau, \alpha}$ where ${\bf 1}_{\phi_{{\tau-1}, {\alpha-1}}}(\phi_{\tau, \alpha})$ is an indicator function that returns 1 if $\phi_{\tau, \alpha}=\phi_{\tau-1, \alpha-1}$.
  
As described in Algorithm \ref{alg:dynamic-programming}, to search for a query with $|q|$ action components in a video with $N$ documents, our dynamic programming approach creates an $N \times |q|$ matrix, $V$, which is filled out from the top left to the bottom right.  A path through the matrix is defined as a set of adjacent (by an 8-neighborhood) matrix element, where each element of the path represents a hypothetical assignment of an action component taking place in a document.  A path containing element $V_{a,b}$ would indicate that action component $b$ occurred in document $a$.

\begin{figure}[h]
    \begin{center}
      \includegraphics[width=.85\linewidth]{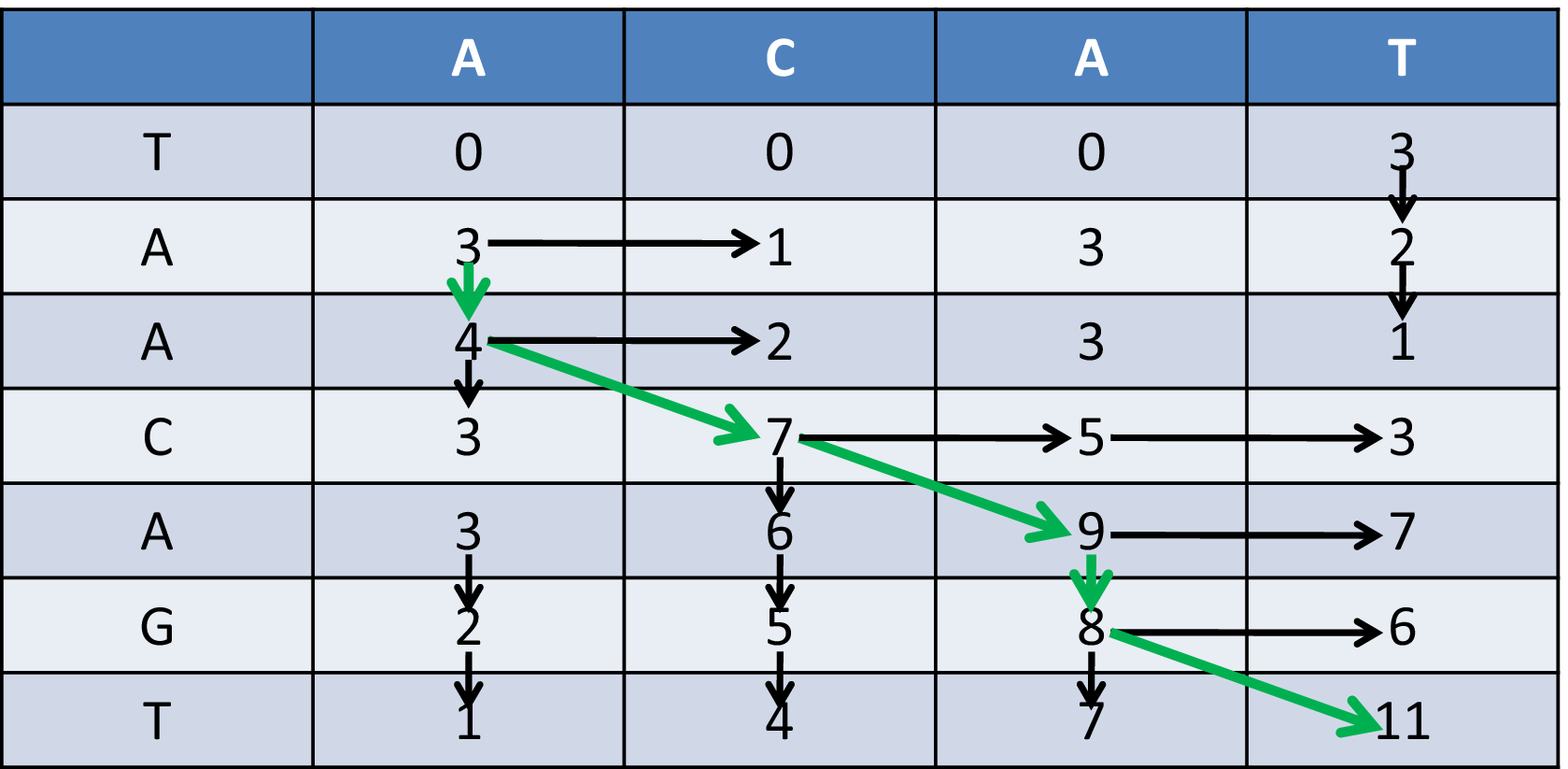}
    \end{center}
    \caption{\small Example of the $V$ matrix.  The query is actions {A, C, A, T}, and the seven documents in the video corpus each contains a single action, {T, A, A, C, A, G, T}.  The values for $W_I, W_D,
  W_C$ and $W_M$ are {$-1, -2, 1 $ and $3$} in this example. The optimal path, {A,A,C,A,G,T}, involves an insertion, a continuation and a deletion.  It is found by tracing backwards from the maximal element, valued at 11.\vspace{-0.1cm}}    
    \label{fig:SWExample}
  \end{figure}

  As the matrix is filled out, each element chooses to append itself to the best possible preceding path - which, by definition, ends immediately above it, immediately to the left of it, or both.  It stores the resulting value, and a pointer to the preceding element, in the value matrix $V$. When the matrix is fully filled out, the optimal path can be found by starting at the maximal value in the matrix and tracing backwards. In order to find multiple matches, we repeatedly find the maximal value in $V$, the optimal path associated with it, set the values along that path to zero, and repeat until the maximum value in $V$ is below a threshold $T$. This threshold will be called the Retrieval Score Threshold for future references. An example of this matrix, with paths overlaid, is shown in figure \ref{fig:SWExample}.

  For a given penalty on each type of distortion $W_I, W_D,
  W_C$ (corresponding to insertion, deletion, and continuation) and a given bonus for each match, $W_M$, the DP
  algorithm (Algorithm \ref{alg:dynamic-programming}) is guaranteed to find the set of partial
  matches which maximizes the sum of the penalties and bonuses over an interval
  of time.  For our queries, we were relatively certain that elements of the query would not be obscured, but we were uncertain about our detection rate on features and how long an event would take.  Thus, we set $W_I=-1$, $W_D=-2$, $W_C=1$, and $W_M=3$. These values preclude optimal solutions which involve deletions, look for longer sequences that match, and are relatively robust to missed detection.  For Airborne footage we also have a bonus $W_{S}$ for a match generated by the same tracklet. In our experiments $W_{S}$ is set to $5$. 

  We note that because it reasons over specific partial matches, our dynamic programming approach
  also finds the locations in the video segments where the event occurs, but
  this is not exploited in the results of this paper.
\vspace{-0.3cm}
\section{Experimental Results}
\label{sec:results}
\vspace{-0.1cm}
  \begin{figure*}[htb!]
  \begin{center}
      \includegraphics[width=0.18\linewidth]{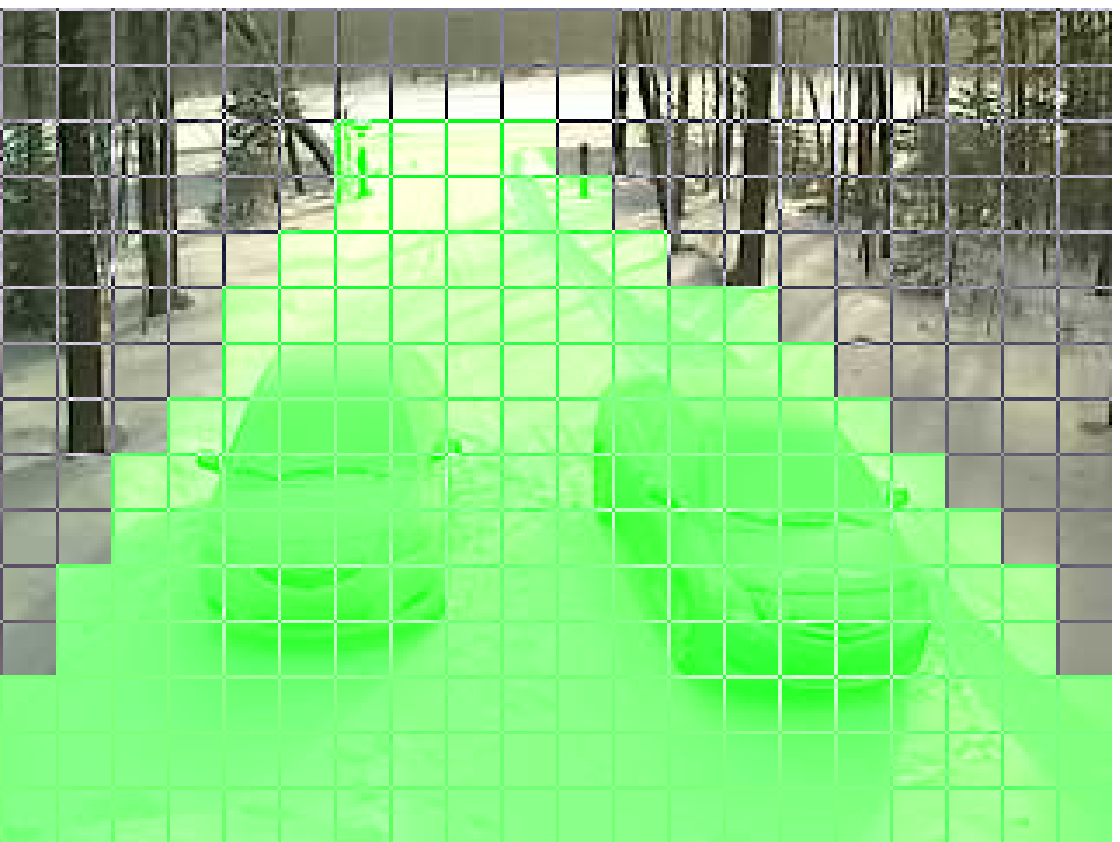}
      \includegraphics[width=0.18\linewidth]{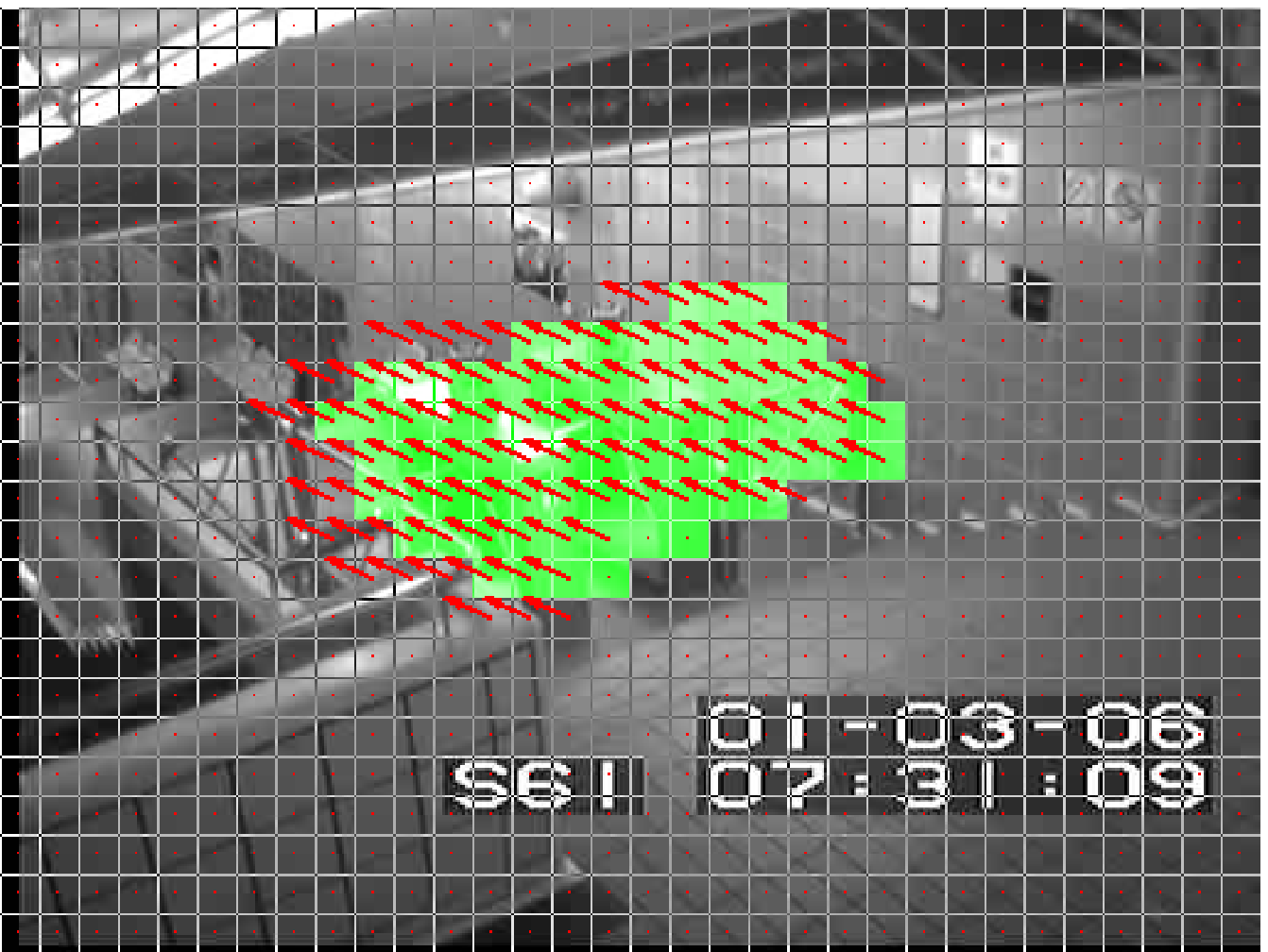}
      \includegraphics[width=0.18\linewidth]{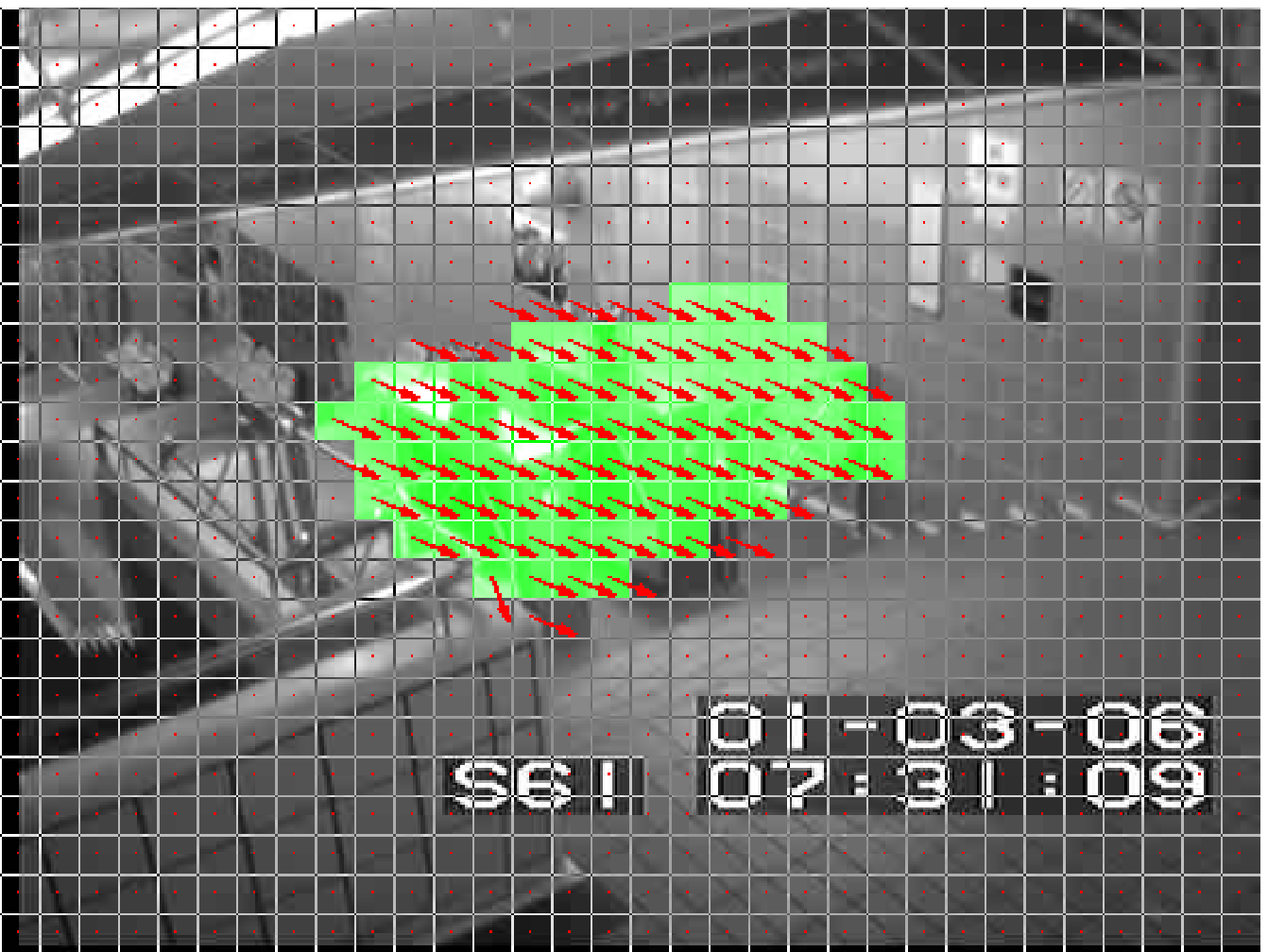}
      \includegraphics[width=0.18\linewidth]{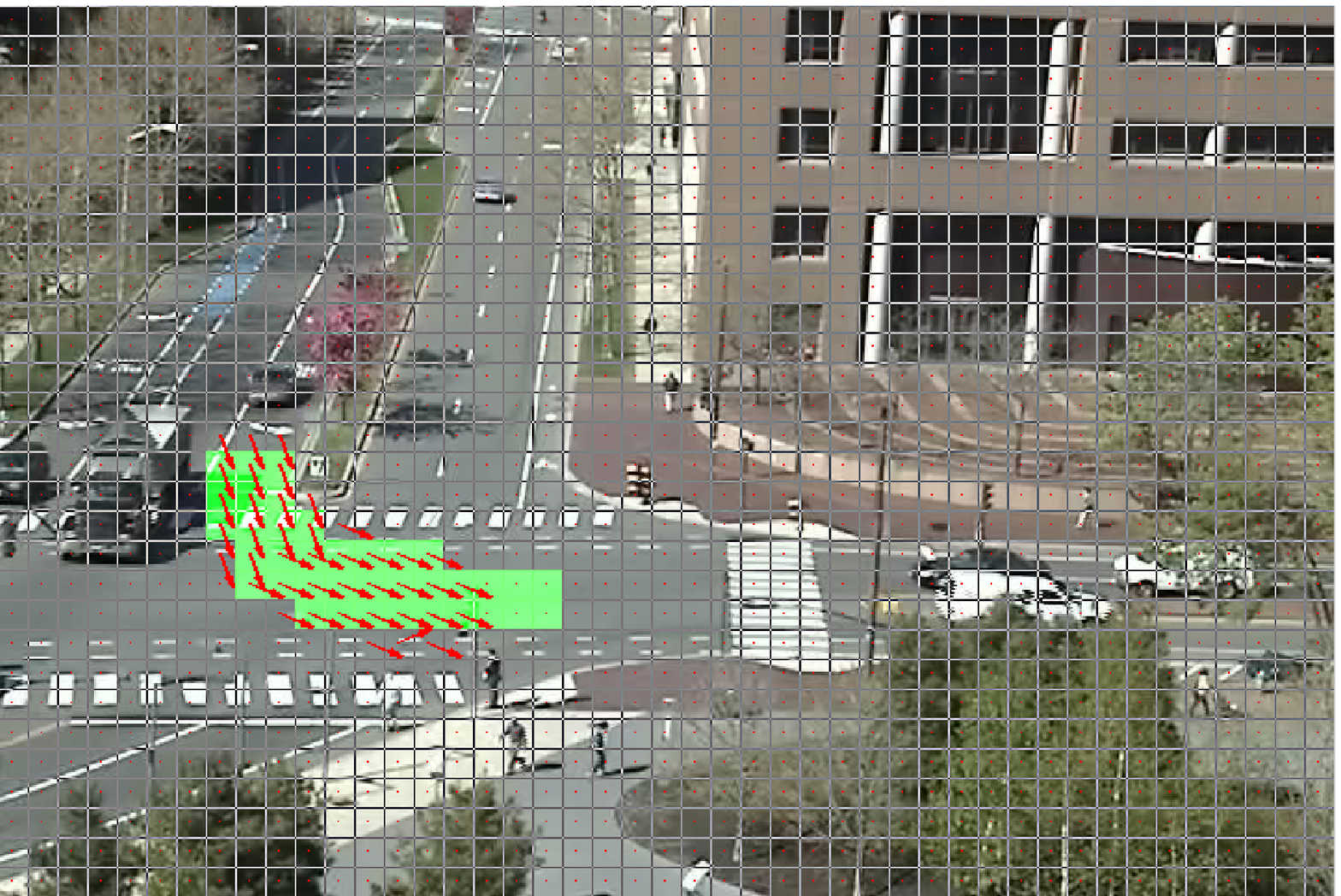}
      \includegraphics[width=0.18\linewidth]{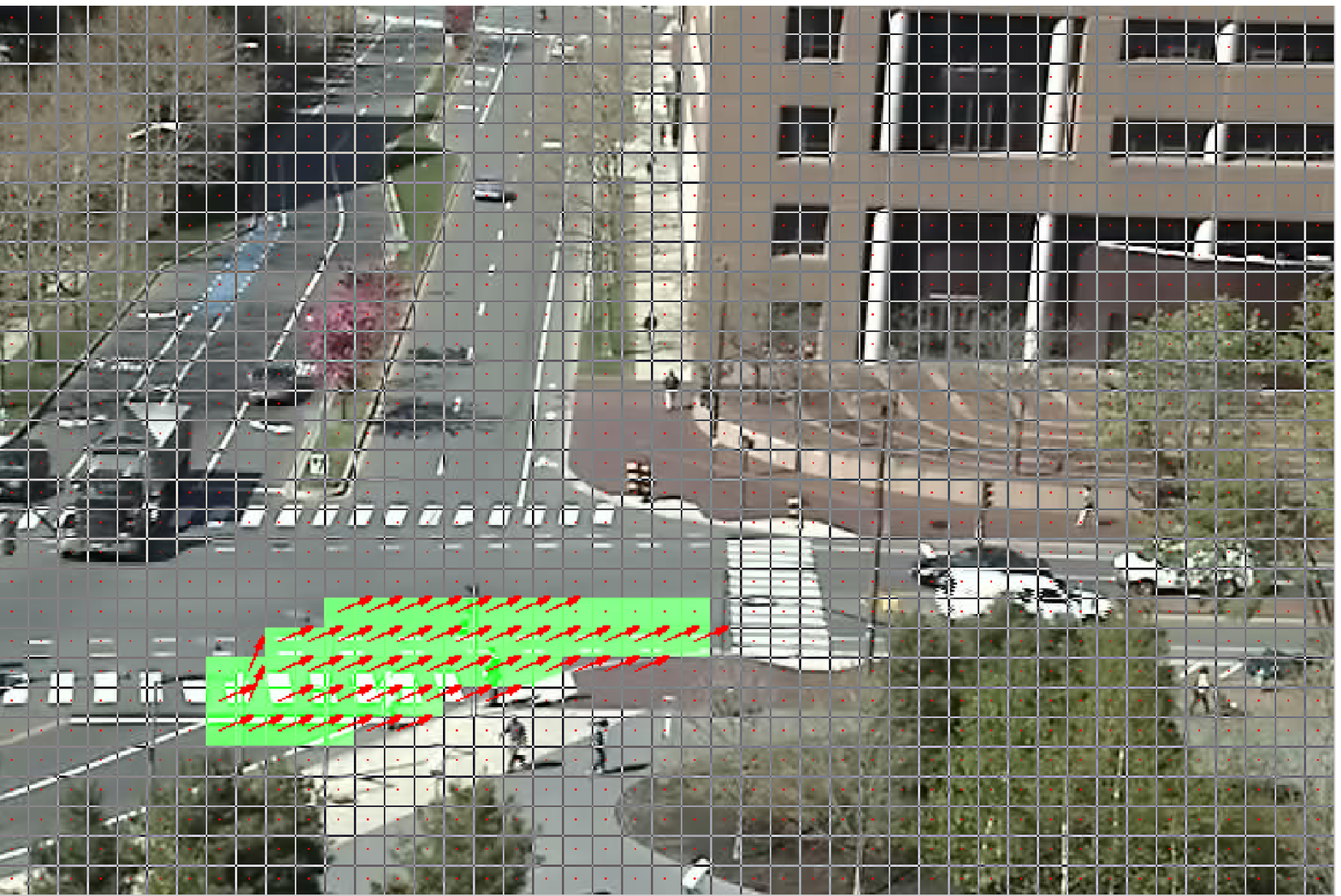}
      \\
      \includegraphics[width=0.18\linewidth]{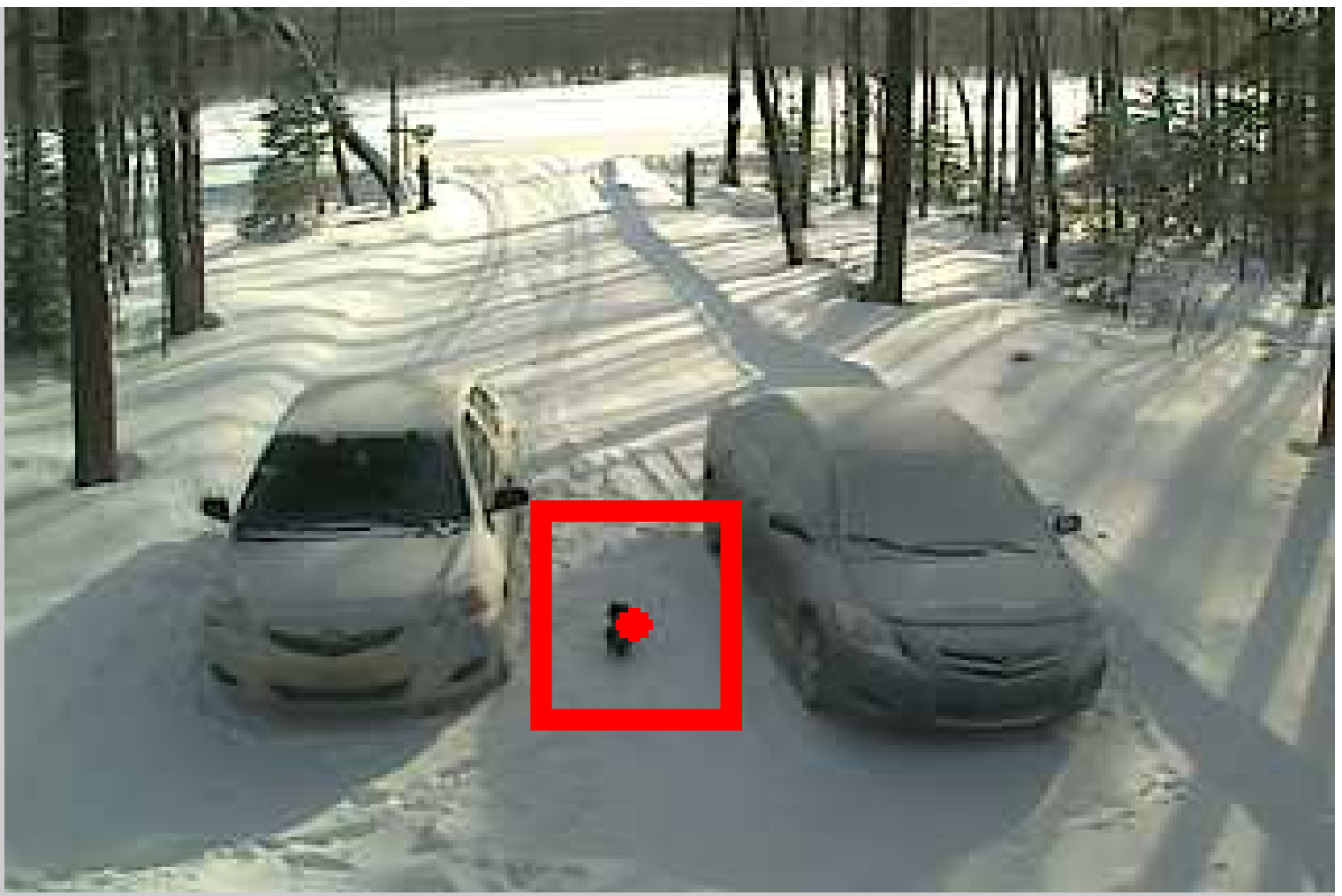}
      \includegraphics[width=0.18\linewidth]{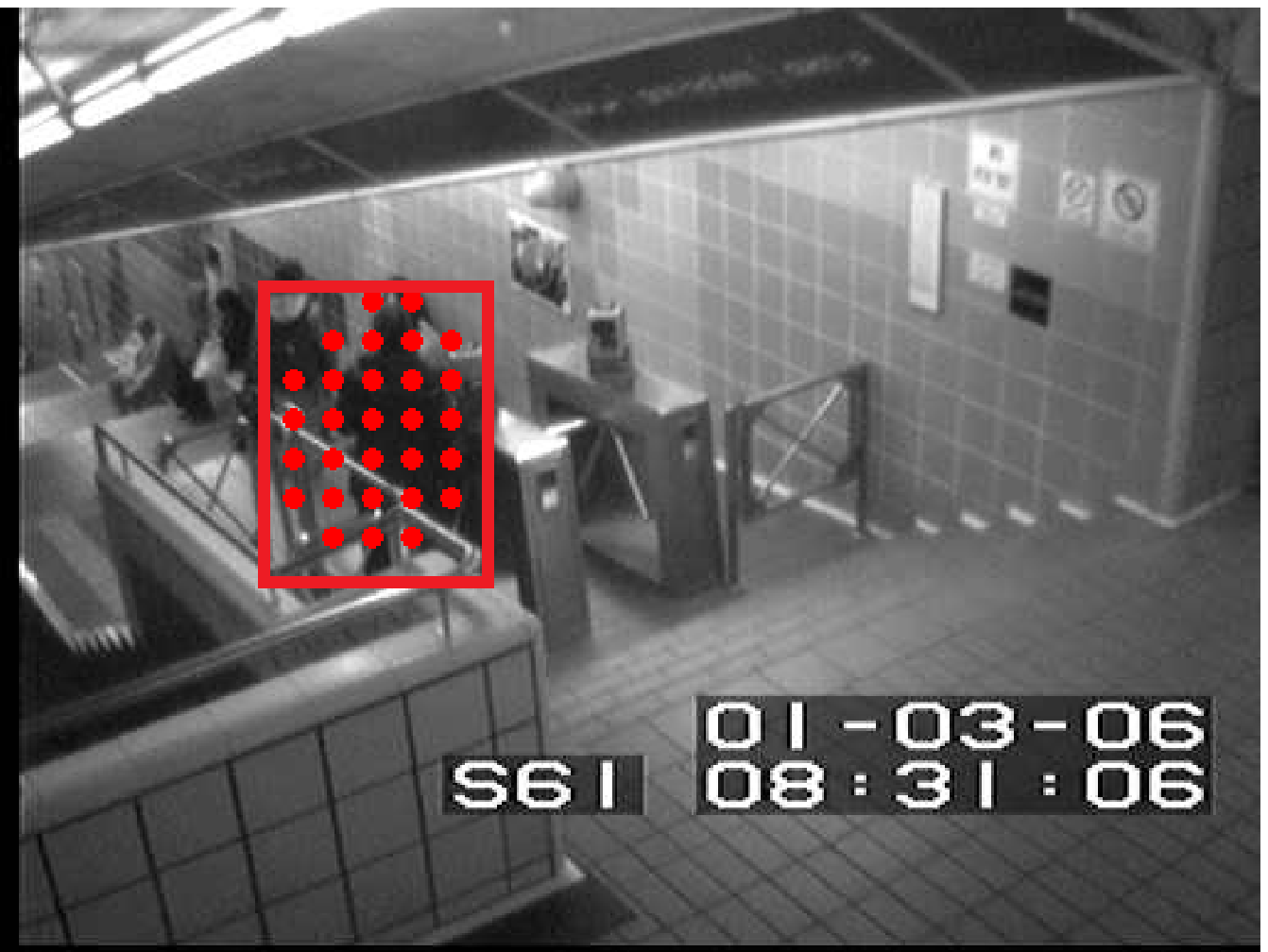}
      \includegraphics[width=0.18\linewidth]{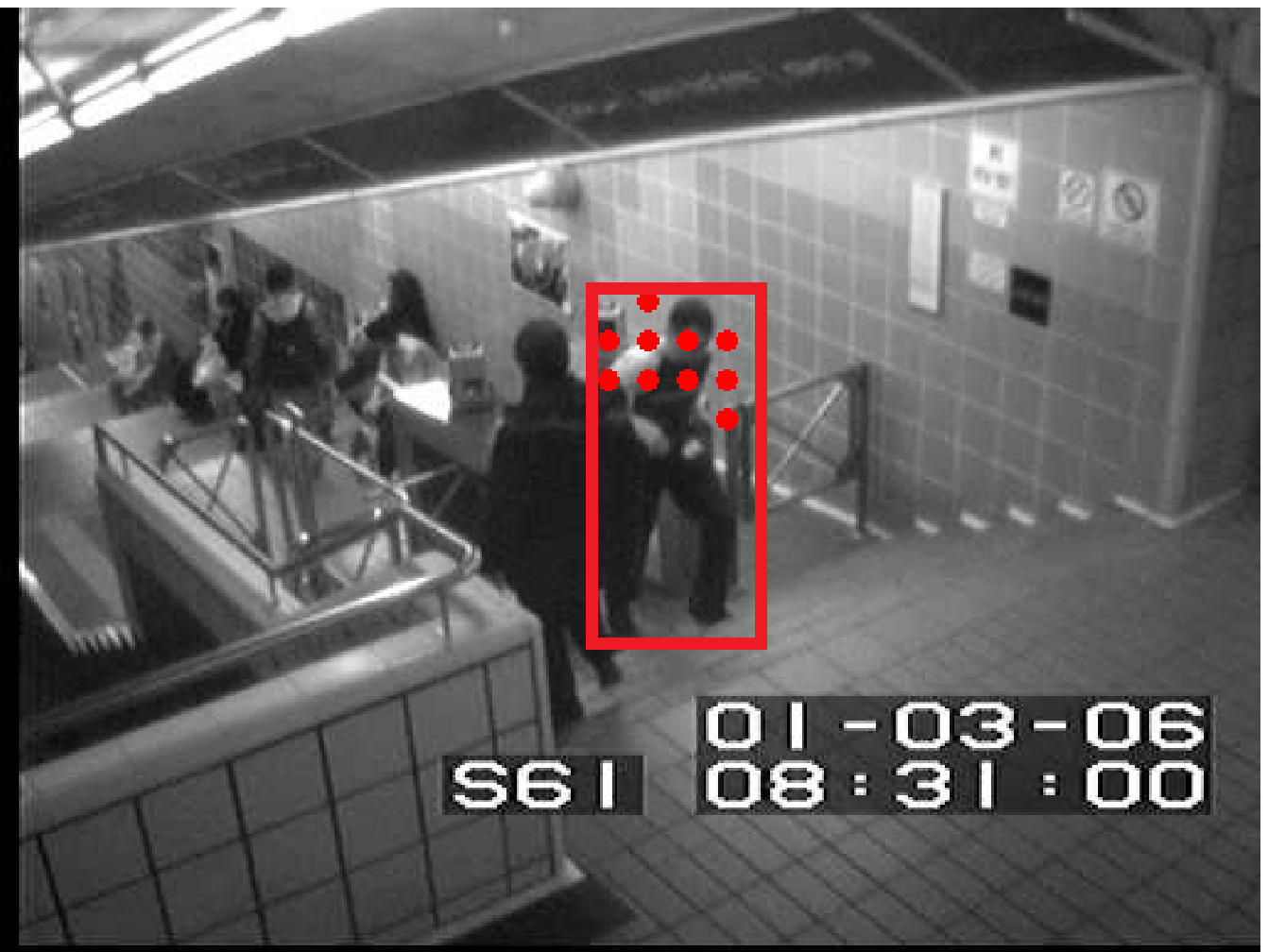}
      \includegraphics[width=0.18\linewidth]{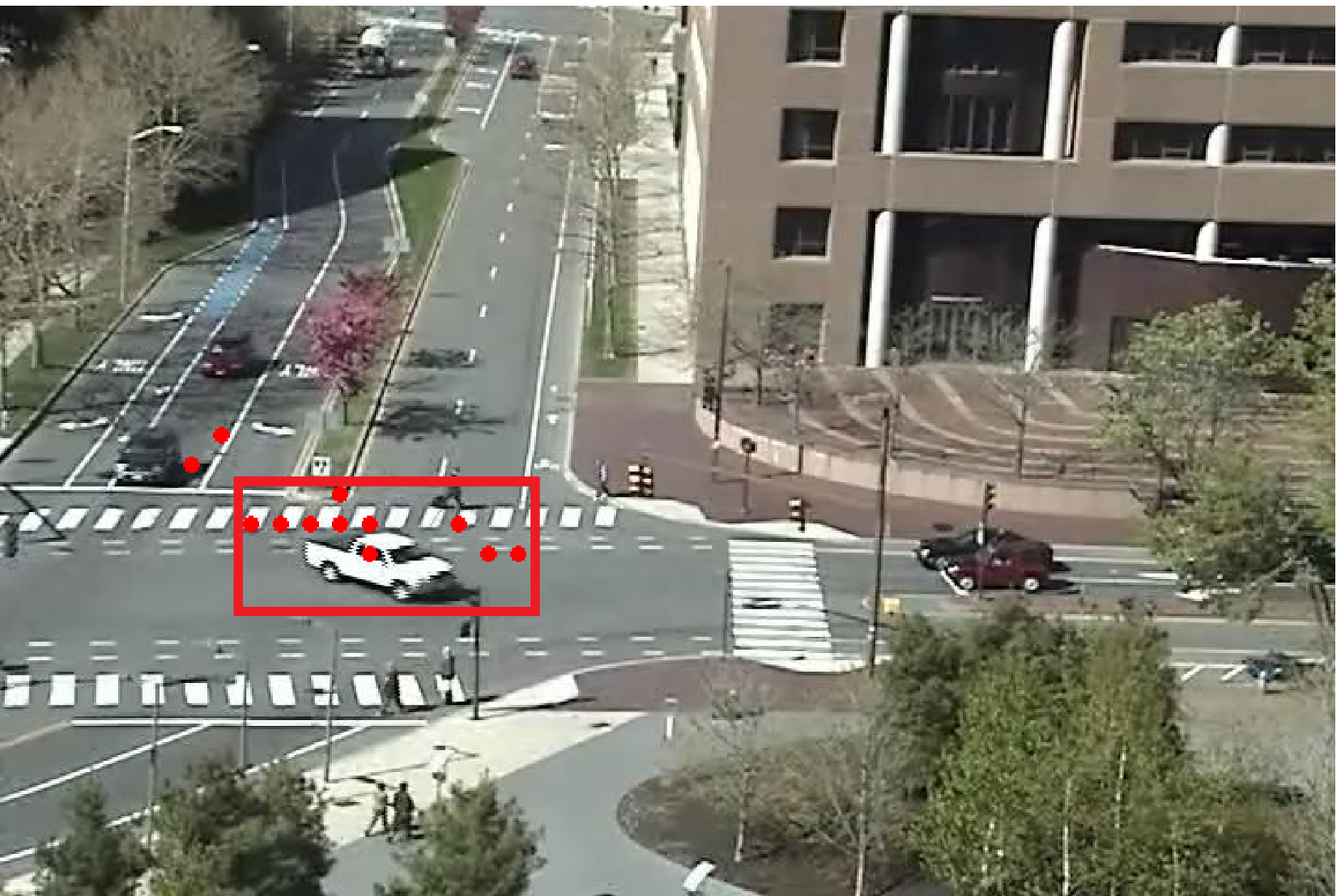}
      \includegraphics[width=0.18\linewidth]{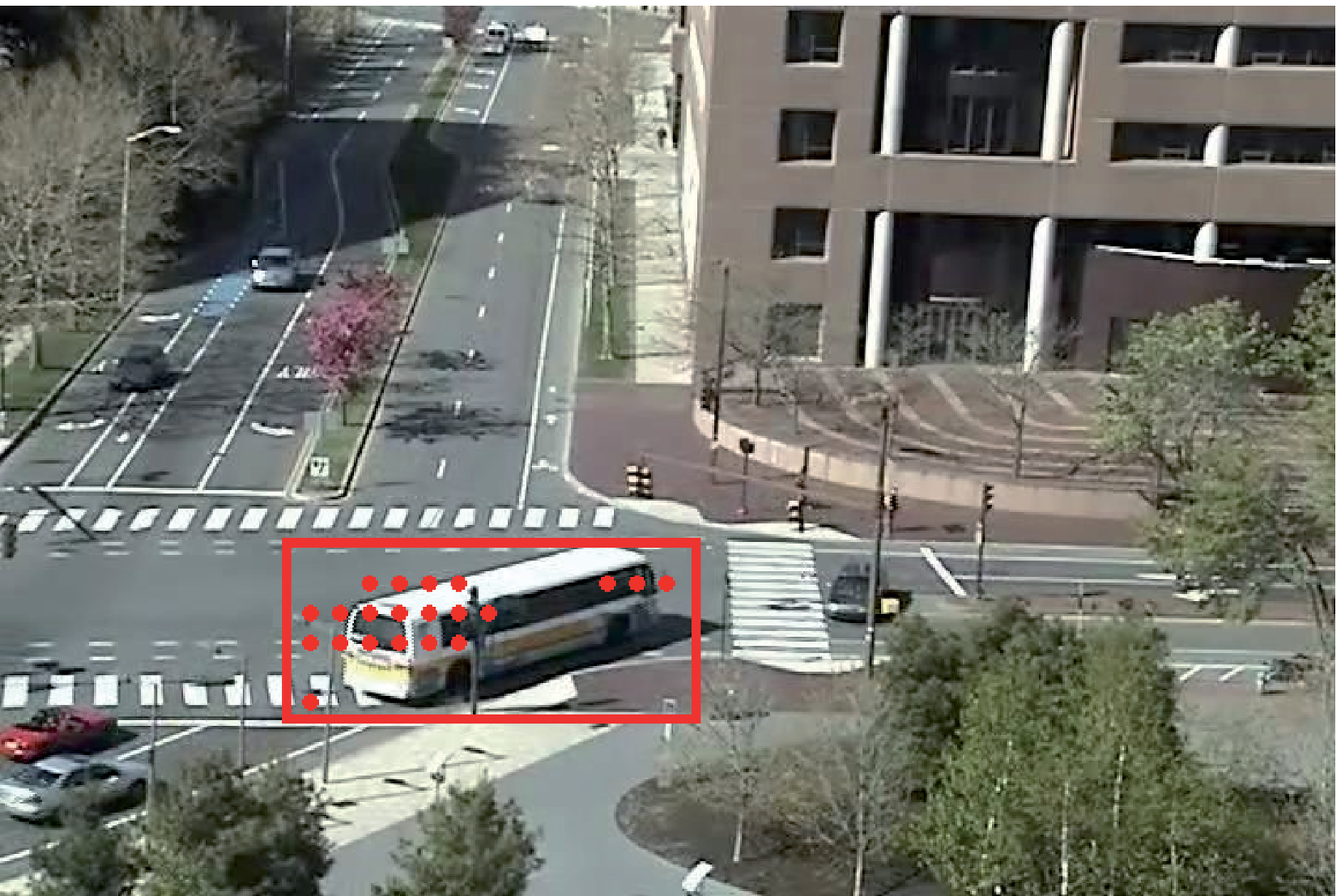}
      \\
      \includegraphics[width=0.18\linewidth]{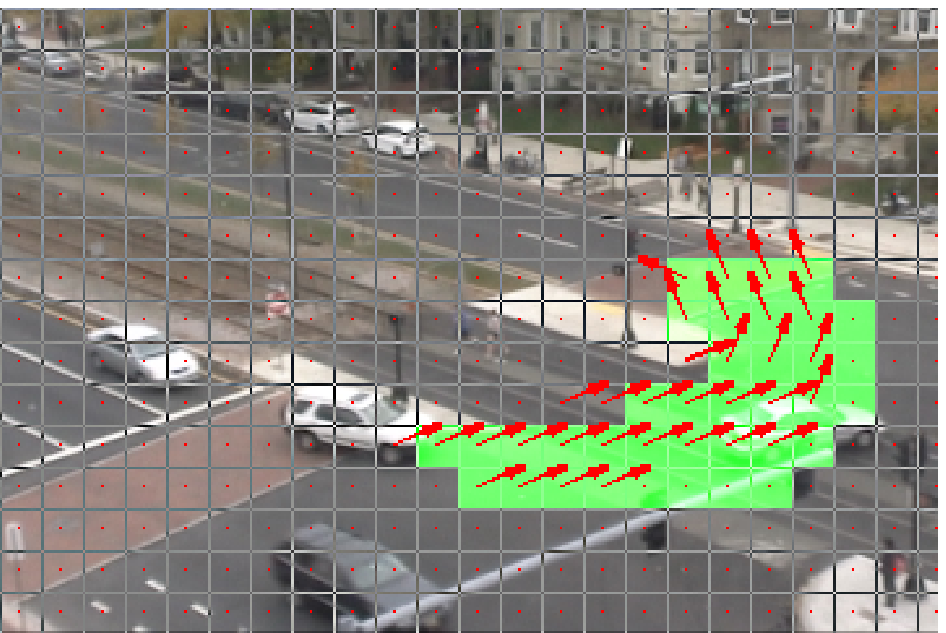}
      \includegraphics[width=0.18\linewidth]{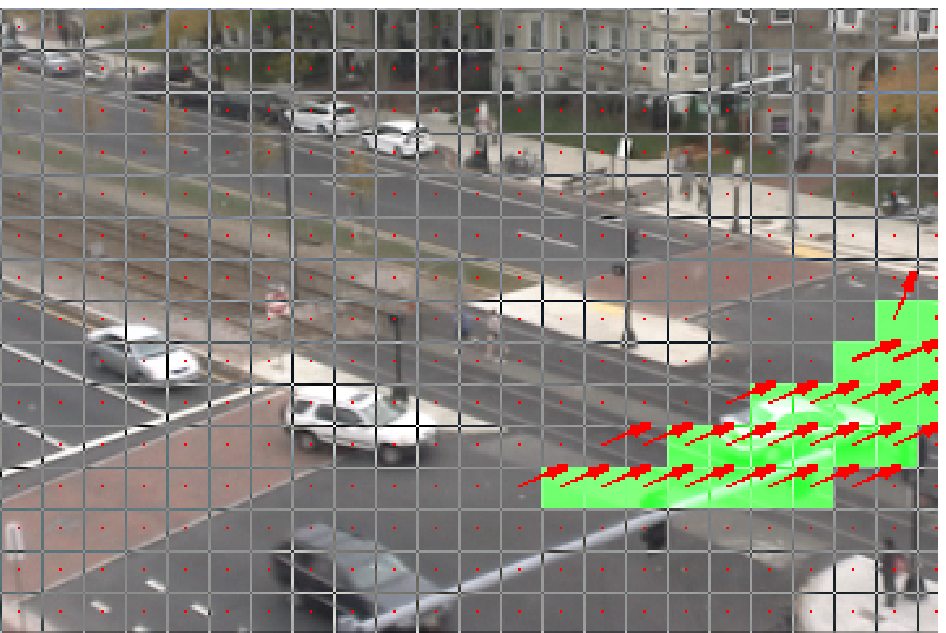}
      \includegraphics[width=0.18\linewidth]{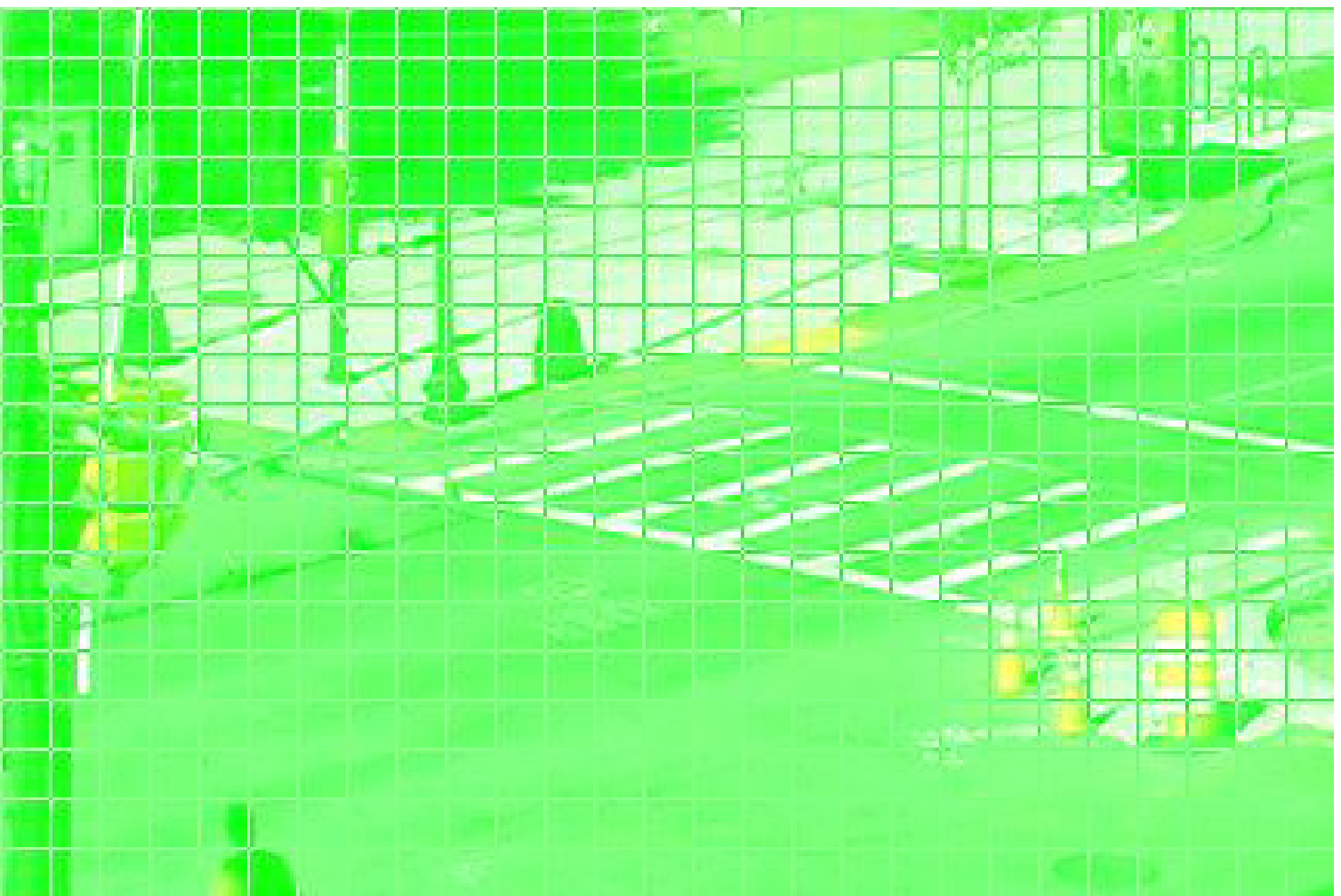}
      \includegraphics[width=0.18\linewidth]{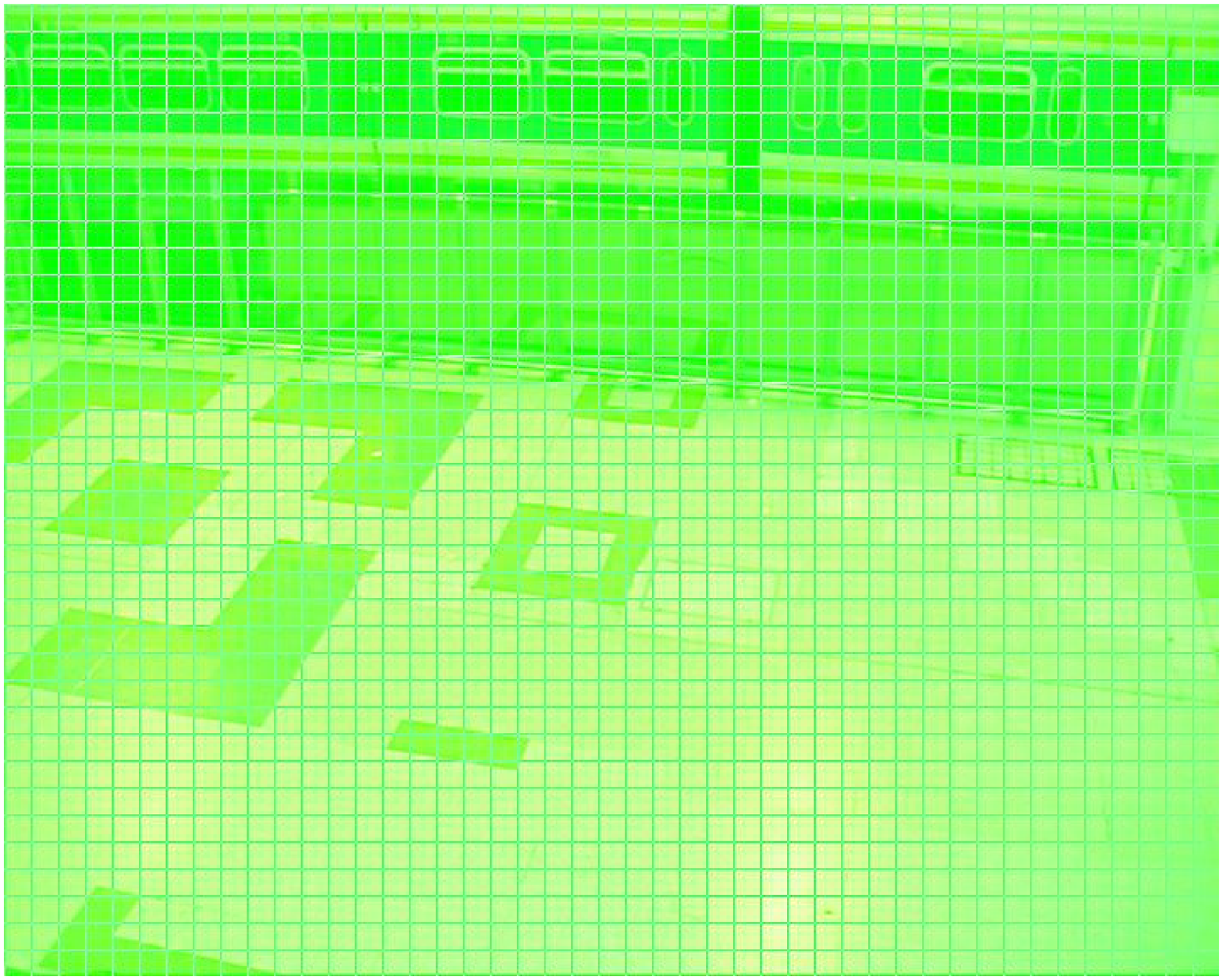}
      \includegraphics[width=0.18\linewidth]{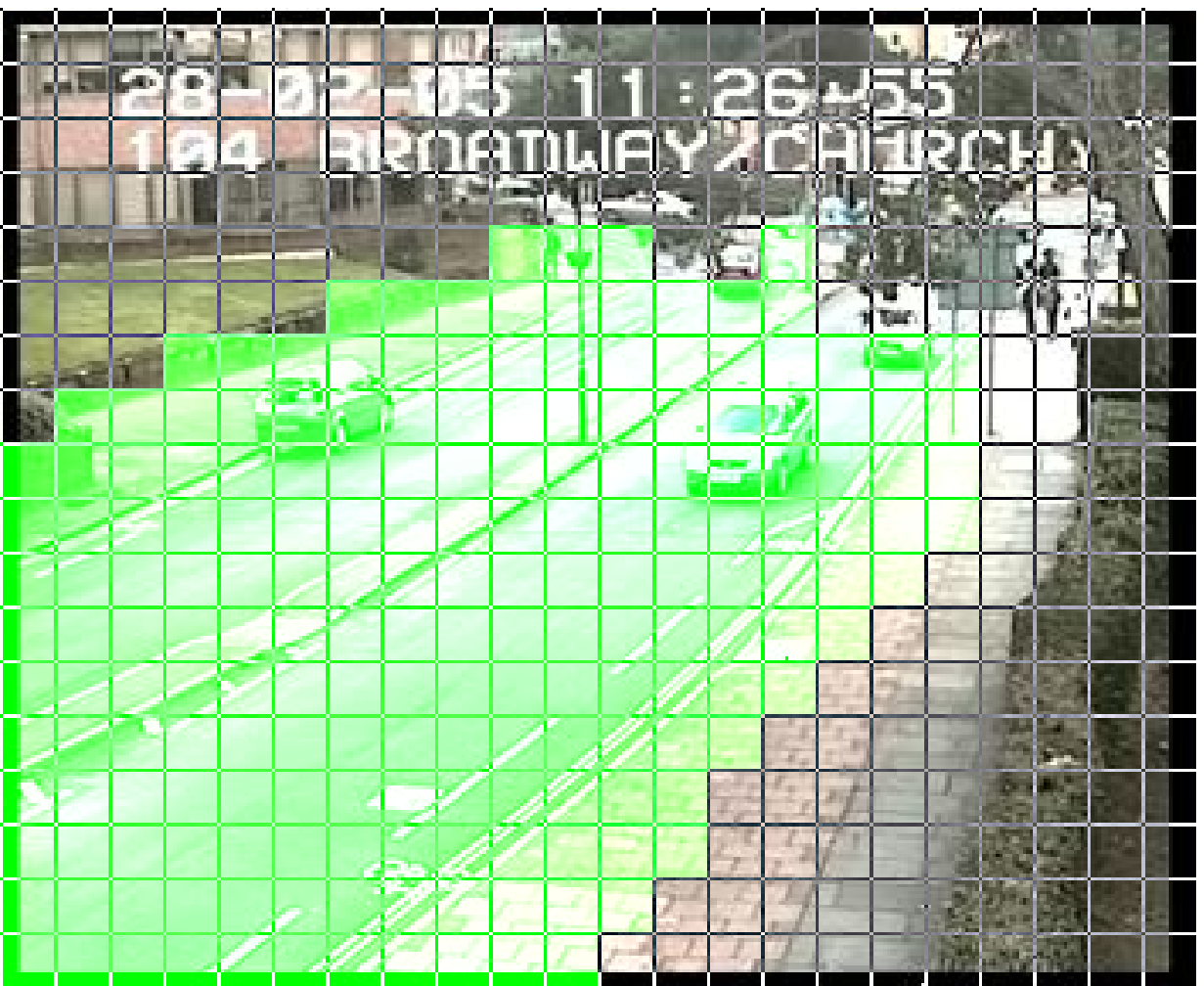}
      \\
      \includegraphics[width=0.18\linewidth]{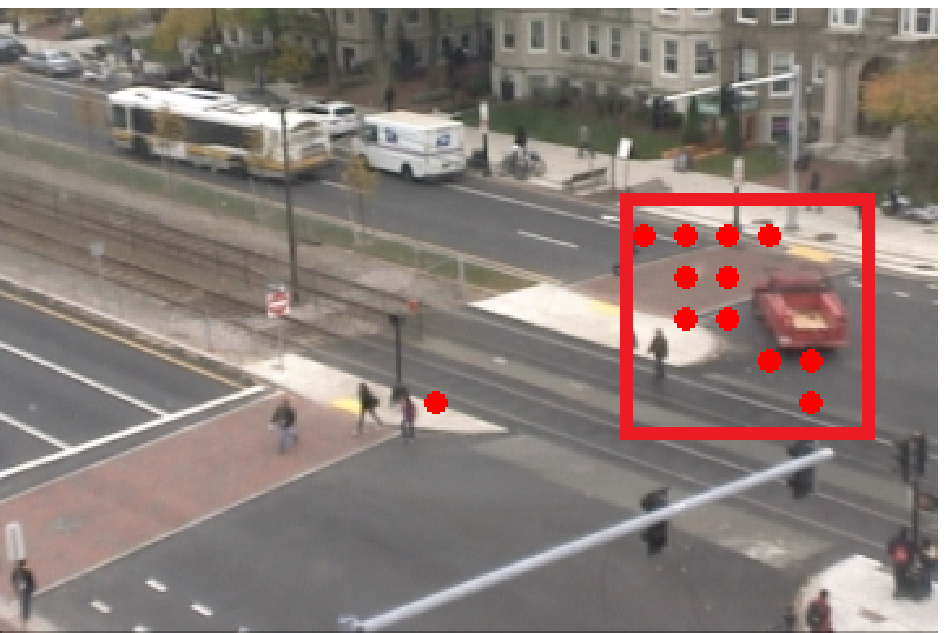}
      \includegraphics[width=0.18\linewidth]{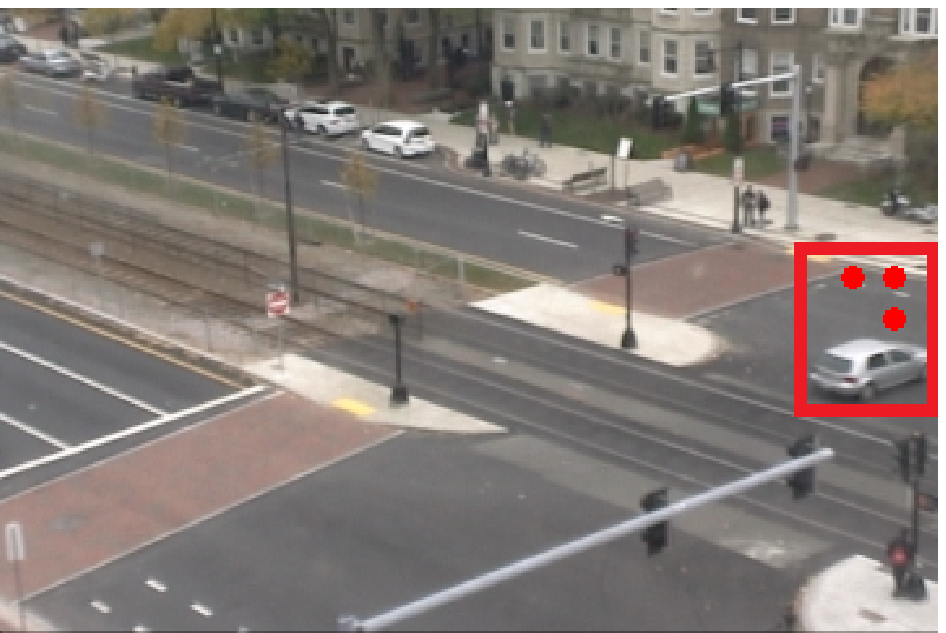}
      \includegraphics[width=0.18\linewidth]{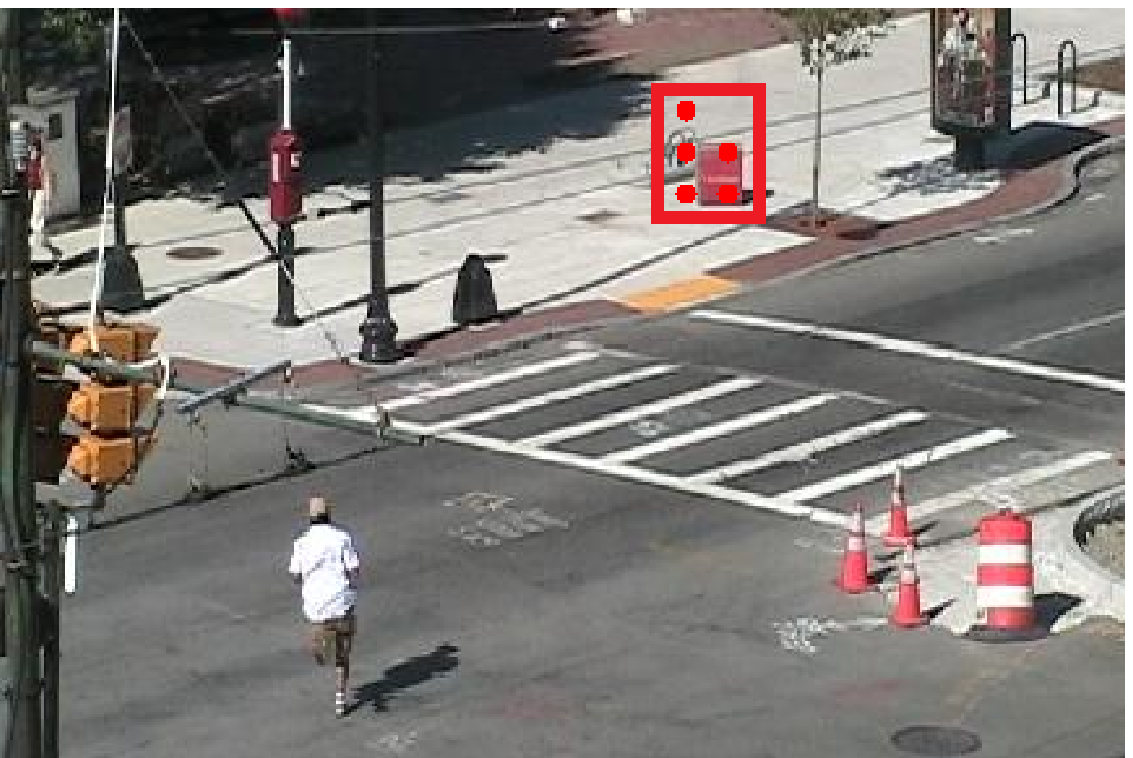}
      \includegraphics[width=0.18\linewidth]{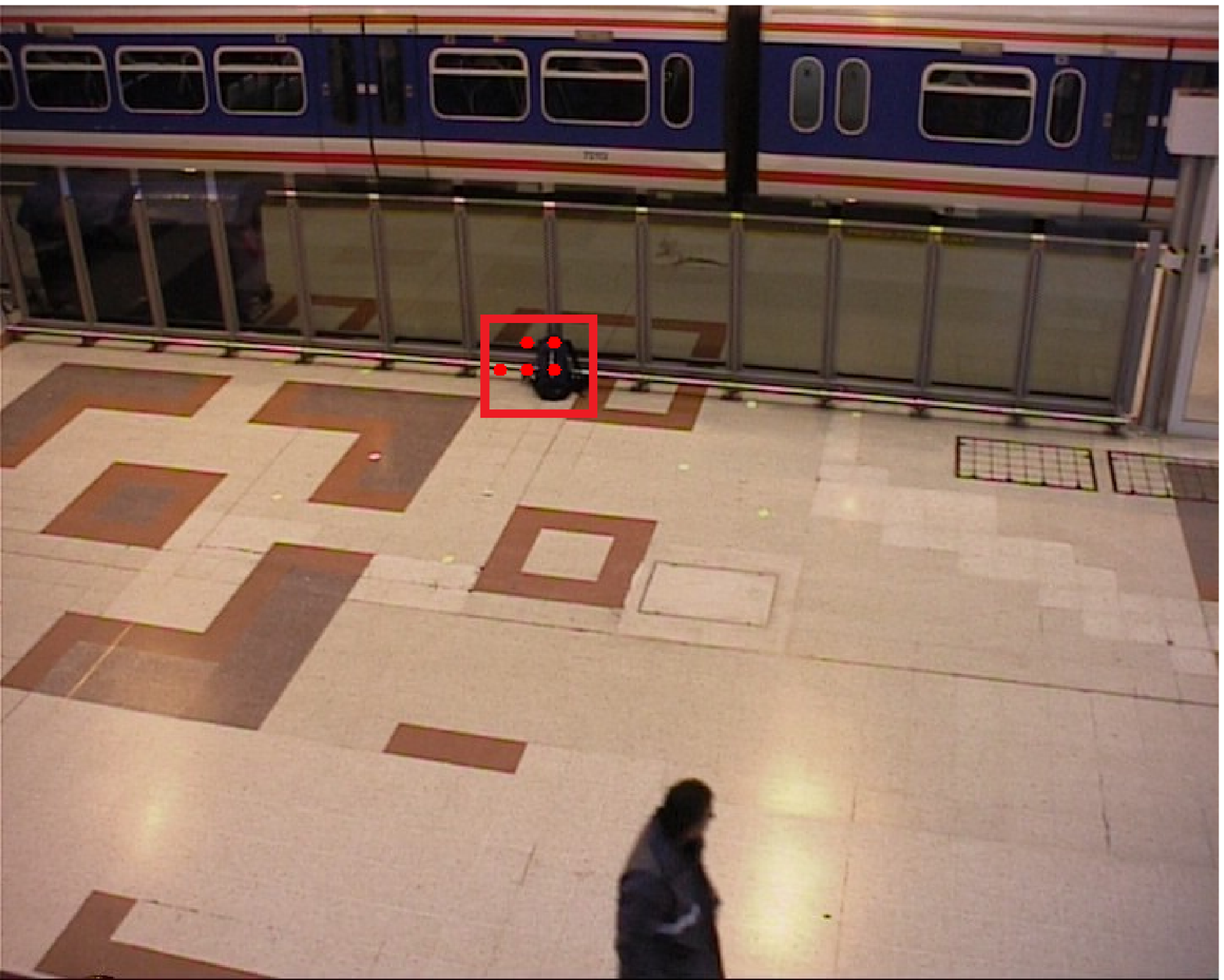}
      \includegraphics[width=0.18\linewidth]{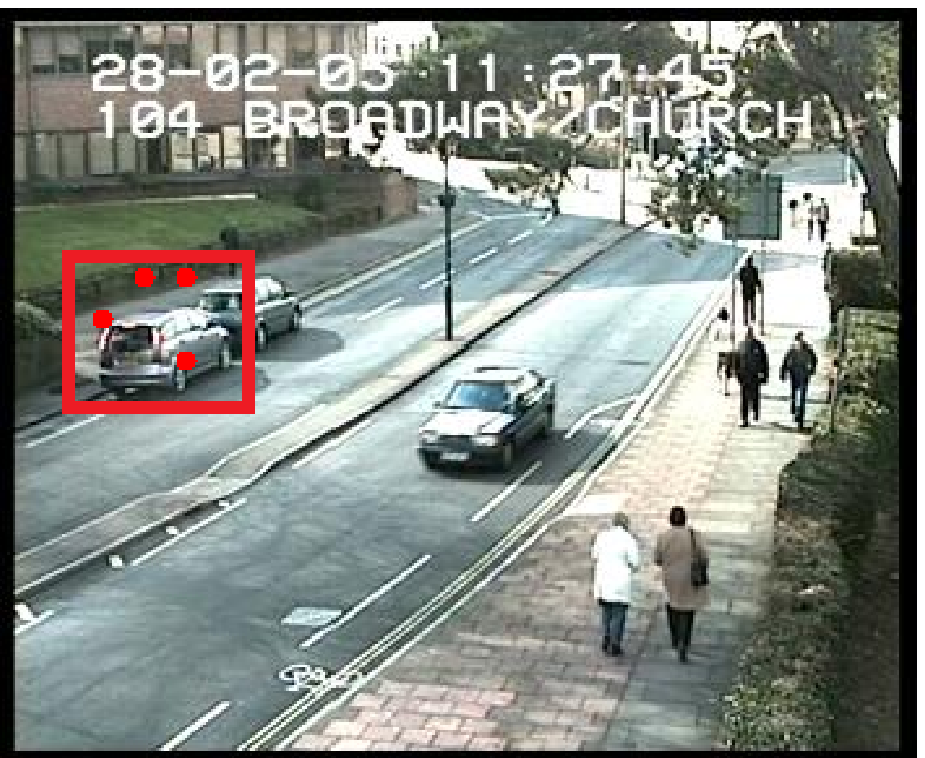}\vspace{-0.4cm}
    \end{center}

    \caption{\small Screen shots of the ten tasks.  These images show the search
      queries (with green ROI) and a retrieve frame (with a red rectangle).  The
      red dots correspond to the tree whose profile fit the query.}
    \label{fig:results}
\end{figure*}


\begin{table*}[htb!]
    \begin{center}
    \begin{tabular}{|l||c|c|c|c|c|c|}

      \hline
      Task & Video & Search query & Features & Duration (min.) & Video size & Index size\\
      \hline
       1 & Winter driveway  &  black cat appearance          & color and size              & 253       & 6.55 GB   & 147 KB \\
       2 & Subway           &  people passing turnstiles     & motion                      & 79        & 2.75 GB   & 2.3 MB\\
       3 & Subway           &  people hopping turnstiles     & motion                      & 79        & 2.75 GB   & 2.3 MB\\
       4 & MIT Traffic      &  cars turning left             & motion                      & 92        & 10.3 GB   & 42 MB\\
       5 & MIT Traffic      &  cars turning right            & motion                      & 92        & 10.3 GB   & 42 MB\\
       6 & U-turn           &  cars making U-turn            & motion                      & 3.4       & 1.97 GB   & 13.7 MB\\
       7 & U-turn           &  cars turning left, no U       & direction                   & 3.4       & 1.97 GB   & 13.7 MB\\
       8 & Abandoned object &  abandoned objects             & size and persistence        & 13.8      & 682 MB    & 2.6 MB\\
       9 & Abandoned object &  abandoned objects             & size, persistence and color & 13.8      & 682 MB    & 2.6 MB\\
      10 & PETS             &  abandoned objects             & size and persistence        & 7.1       & 1.01 GB   & 5.63 KB\\
      11 & Parked-vehicle   &  parked vehicles               & size and persistence        & 32        &           &\\
      12 & Intersection 1   &  cars moving                   & tracklets                  & 13.8      & 1.16 GB   &  153 KB \\
      13 & Beach 1          &  people walking                & tracklets                  & 13.8      & 529 MB    &  65 KB \\
      \hline
    \end{tabular}
    \end{center}
    \caption{\small Tasks' number, videos, search query, associate
      features, video duration, video size and index size.  Videos of Task 12 and 13 have a frame rate of 2 frames per second. Tasks 1, 8, 9, 10, 11, 12 and 13 use compound search operators. The index size can be several orders of magnitude smaller than raw video. Our use of primitive local features implies that index times and index size are both proportional to the number of foreground objects in the video. Consequently, index size tends to be a good surrogate for indexing times.\vspace{-0.8cm}}
    \label{tab:task1}
\end{table*}

  \subsection{Datasets}
  \label{sec:datasets}
  \vspace{-0.1cm}	
    In order to evaluate performance of the two-step problem formulation, and the
  DP approach in particular, we tested our two-step approach on eleven CCTV
  videos and two Airborne videos (see table~\ref{tab:task1}, Fig.~\ref{fig:results} and Fig.~\ref{fig:RoutesExamples}).  These videos were selected to test the application of this basic approach to multiple domains, as well as to provide a basis for comparison to other algorithms.  The {\em Winter driveway}, {\em U-Turn} and {\em Abandoned object} sequences were shot by
  us, {\em PETS} and {\em Parked-vehicle} and {\em MIT-traffic} come from known
  databases~\cite{PETS06,ILIDS}, {\em MIT-traffic} was made available to us by
  Wang {\em et al.}~\cite{MIT}; {\em Subway} from Adam {\em et
    al.}~\cite{Adam08}. {\em Airborne} was made available to us by our sponsors.

	As listed in Table~\ref{tab:task1}, we tested different queries to recover
  moving objects based on their color, size, direction, activity, persistence and tracklets.  We
  queried for rare and sometimes anomalous events (cat in the snow, illegal
  U-turns, abandoned objects and people passing turnstile in reverse) as well as
  usual events (pedestrian counting, car turning at a street
  light, and car parking).  Some videos featured events at a distance {\em MIT-traffic}, while others featured people moving close to the camera {\em Subway}.  We searched for objects, animals, people, and vehicles moving along user-supplied routes. 

\vspace{-0.3cm}	
  \subsection{Examined Tasks}
  \label{sec:eleventasks}
\vspace{-0.1cm}	

\begin{figure}
    \begin{center}
      \includegraphics[width=0.99\linewidth]{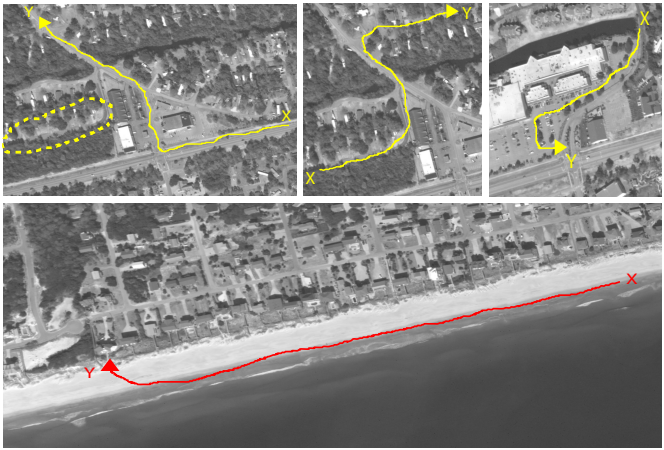}\vspace{-0.3cm}
    \end{center}
    \caption{\small Examples of the examined routes. Routes are shown in yellow/red arrows and they start from point X and end at point Y. Some of the routes undergo strong occlusion (see dashed yellow region, top row, first column, for route in second column) and others undergo many turns (see first row, second and last column).\vspace{-0.2cm}} 
    \label{fig:RoutesExamples}
  \end{figure}

  We examined 11 CCTV footage (see Fig.\ref{fig:results} for examples) and 2 Airborne sequences (Fig.\ref{fig:RoutesExamples}). Once we had defined the queries, manually created a ground-truth list for each task consisting of ranges of frames.  Comparison is obtained by computing the
  intersection of the ranges of frames returned by the search procedure to the
  range of frames in the ground truth.  An event is marked as detected if it appears in the
  output video and at least 1 partial match hits objects appearing in the event. For airborne footage, we also require that the start and end frames be within 15 frames of the ground truth.
	
	\begin{table*}
    \begin{center}
    \begin{tabular}{|l||c|c|c|c|c|c|c|c|}

      \hline
      Task & Video & Ground Truth & Greedy True    & HDP~\cite{Kuettel10} True &Greedy False     & HDP~\cite{Kuettel10} False  & Runtime  \\
           &       & (events)     & (events found) & (events found)            &(events found)   & (events found)              & (seconds)\\
      \hline
       1 & Winter driveway  & 3   & 2   & -- & 1  & --  &  7.5 \\
       2 & Subway           & 117 & 116 & 114& 1  & 121 &  0.3 \\
       3 & Subway           & 13  & 11  & 1  & 2  & 33  &  3.0 \\
       4 & MIT Traffic      & 66  & 61  & 6  & 5  & 58  &  0.4 \\
       5 & MIT Traffic      & 148 & 135 & 54 & 13 & 118 &  0.5 \\
       6 & U-turn           & 8   & 8   & 6  & 0  & 23  &  1.2 \\
       7 & U-turn           & 6   & 5   & 4  & 1  & 14  &  0.6 \\
       8 & Abandoned object & 2   & 2   & -- & 0  & --  &  4.8 \\
       9 & Abandoned object & 2   & 2   & -- & 0  & --  & 13.3 \\
      10 & PETS             & 4   & 4   & -- & 0  & --  & 20.2 \\
      11 & Parked-vehicle   & 14  & 14  & -- & 0  & --  & 12.3 \\
      \hline
    \end{tabular}
    \end{center}
    \caption{\small Results for the elevens tasks using greedy optimization and HDP labels.  Crossed-out
      rows correspond to queries for which there was no corresponding topic in the HDP search.\vspace{-0.7cm}}
    \label{tab:task2}
\end{table*}

	\vspace{-0.2cm}
	\subsection{Comparison with other techniques}
	\vspace{-0.2cm}
	\label{sec:OtherTechniques}
	
	For purposes of comparison we implemented two other approaches: the greedy approach described in section \ref{sec:greedy-algorithm}, and a scene understanding technique based on Hierarchical Dirichlet Processes ~\cite{Xiang08,Kuettel10} described in section \ref{sec:HDP}.  For Airborne data, there is also a question of the quality of the tracklet features.  As such, we implement our Dynamic Programming approach with a range of features: 1) tracklets from our algorithm, 2) frame to frame motion on the original Airborne footage (ME+DP), 3) frame to frame motion on binary detection masks (MEB+DP), and KLT trajectories (KLT+DP)\cite{Shi94}. We also compare against explicit Track Matching. For a query route, we estimate the common distance between all available tracklets. A tracklet is a match if the common distance is greater than a certain threshold.

	\subsubsection{HDP Comparison}
	\label{sec:HDP}
	
	For the purposes of comparison, we employed high-level search functions based
  on scene understanding techniques using Hierarchical Dirichelet Processes
  (HDP) ~\cite{Xiang08,Kuettel10}.   At each iteration, the HDP-based learning
  algorithm assigns each document to one or more high-level activities.  This
  classification is used as input to the next training iteration.  Xiang {\it et
    al.} ~\cite{Xiang08} propose a search algorithm that uses learned topics as
  high-level semantic queries.  The search algorithm is based on the
  classification outputs from the final HDP training iteration.  We compare our
  method to this HDP-based search algorithm.

  Queries are specified as the ideal classification distribution and the search
  algorithm compares each document distribution over the learned topics
  against this ideal distribution.  Comparison is performed using the relative
  entropy (Kullback-Leibler divergence) between the two distributions.  The
  Kullback-Leibler divergence gives a measure of distance between the query $q$
  and the distribution $p_j$ for document $j$ over the $K$ topics:
  \begin{eqnarray*}
    D(q,p_j) = \sum_{k=1}^{K}q(k)\log \frac{q(k)}{p_j(k)}.
  \end{eqnarray*}
Query $q$ is created by looking at the ideal documents and assigning to $q$ a uniform distribution over the topics present in them. The search evaluates $D(q,p_j)$ for each document $j$ and ranks the documents in order of increasing divergence.

\vspace{-0.1cm}
\subsection{CCTV Results}
\vspace{-0.1cm}
	
	Results from our greedy-optimization method are displayed in Fig. ~\ref{fig:results} and
  summarized in table ~\ref{tab:task2}.  The comparable results for HDP are
  summarized below in table ~\ref{tab:task2}. The ``Ground truth'' column of table \ref{tab:task2} indicates the true number of events which exist in the dataset.  The ``Greedy True'' column indicates the number of correct detections (true positives) for our Greedy algorithm and ``HDP True'' the number of correct detections (true positives) for the HDP-based search ~\cite{Wang09}.  Likewise, the ``Greedy False'' and ``HDP False'' indicate the number of false alarms that were found for those eleven tasks.

    Table ~\ref{tab:task2} demonstrates the robustness of the two-step method in a
  wide-array of search applications, outperforming the HDP baseline
   in detection and false alarm rate.  The figures in the table are given for
  results of total length approximately equal to that of the ground truth.  As
  can be seen from the figures in the table, the absolute detection rate is
  strong.

  From table ~\ref{tab:task2}, one can see that HDP-search deals well with search of
  recurring large-scale activities and poorly otherwise.  While several queries
  could not be executed because of a lack of topics that could be used to model
  the query, the results nonetheless demonstrate some of the shortcomings of the
  algorithm.  The HDP search scales linearly with the number of documents, an
  undesirable quality with large datasets.  Further, the cost of the training
  phase is prohibitive (approximately 2 days for the ``subway'' sequence) and
  must be paid again every time that more video data is included.

  \subsubsection{Dynamic Programming}
  
\begin{figure*}
    \begin{center}
      \includegraphics[width=0.28\linewidth]{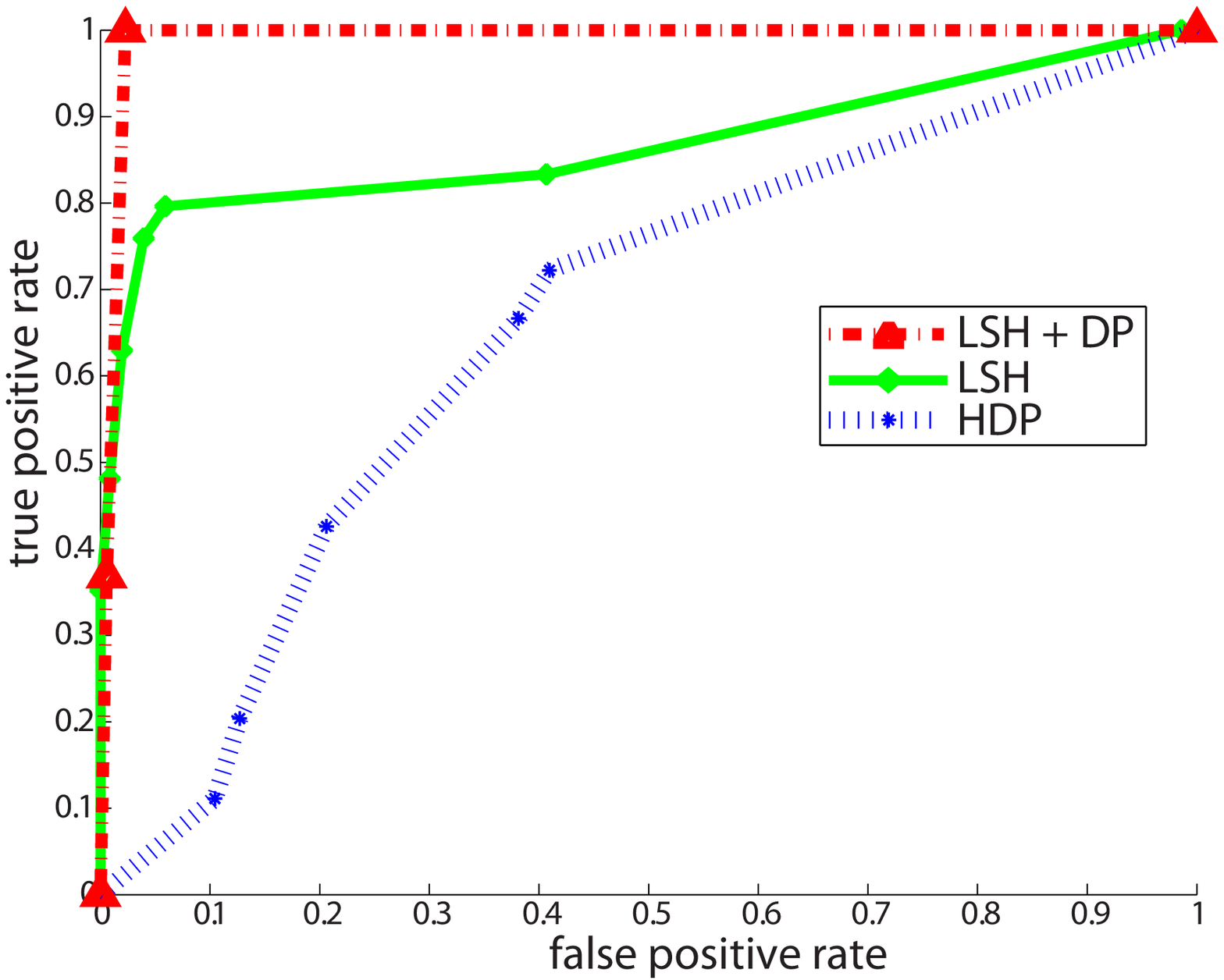}
      \includegraphics[width=0.28\linewidth]{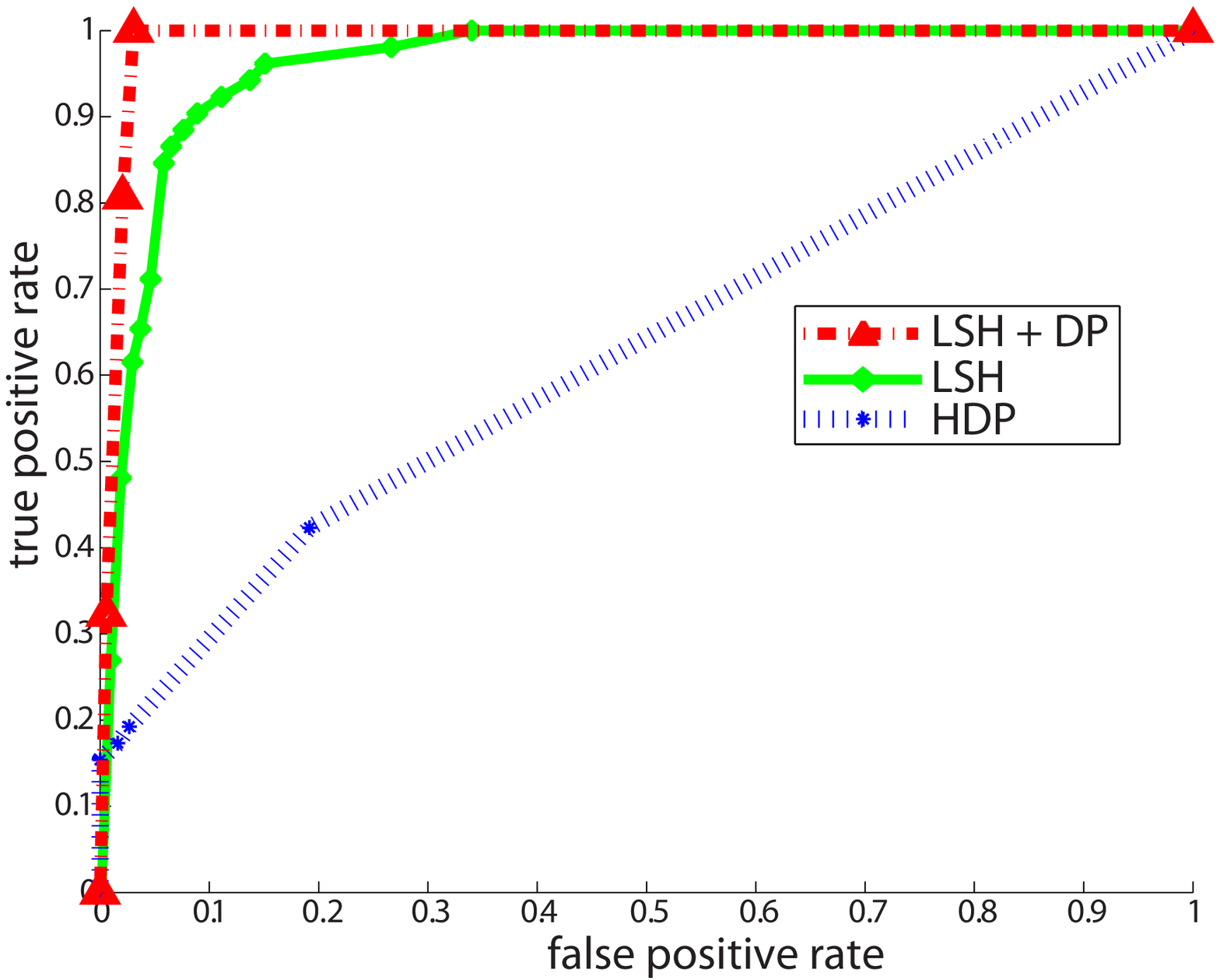}
      \includegraphics[width=0.28\linewidth]{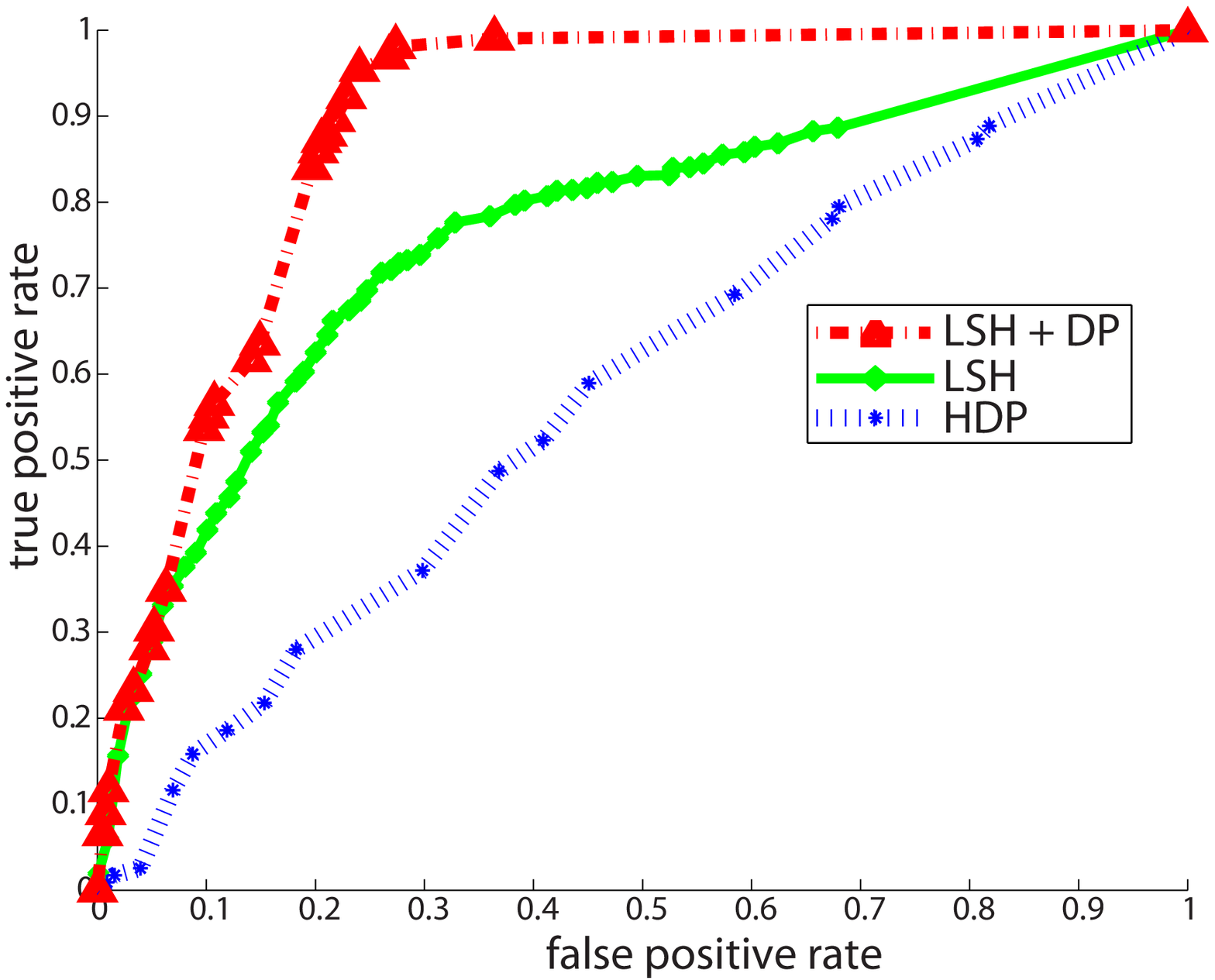}
    \end{center}
    \caption{\small ROC curves for the U-turn, subway and traffic datasets. Both our greedy search and DP significantly outperform scene-understanding methods such as HDP methods \cite{Xiang08,Kuettel10}.\vspace{-0.5cm}}
    \label{fig:roc-curves}
  \end{figure*}

  Two of the seven tasks in table \ref{tab:task1} had temporal structure which could be exploited through dynamic programming.  In order to demonstrate the potential gain from exploiting this structure, we chose tasks three, four and six and performed dynamic programming using the full query, as well as greedy and HDP search algorithms. The ROC curves for those three scenarios are provided in Fig \ref{fig:roc-curves}, contrasting dynamic programming with HDP and Greedy Optimization.

  Fig. ~\ref{fig:roc-curves} demonstrates the type of improvement that
  can be attained by the two-step approach to search.  The ROC curves for
  LSH-based greedy optimization dominate the HDP curves, and there is clear
  improvement from employing time-ordering with DP.  These improvements come as
  no surprise - HDP is doing a global search, attempting to create a topic for each action.  LSH does a compelling local search which is fast and produces low false alarm rate.  This is largely due to the global nature of HDP - the topics it discovers are more likely to be common events, so infrequent events such as u-turns and turnstile-hopping pose a problem.  Note that the gap between local and global searches narrow on the MIT-traffic dataset, where the event being found (left turns at a light) is a relatively common occurrence.

\vspace{-0.1cm}
\subsection{Airborne Results}
\vspace{-0.1cm}

\begin{figure}[tp]
    \begin{center}
      \includegraphics[width=0.99\linewidth]{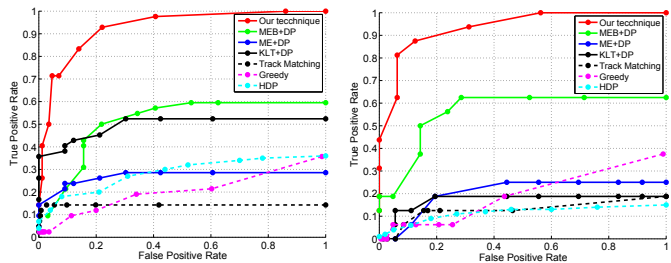}\vspace{-0.3cm}
    \end{center}
    \caption{\small ROC for the examined airborne data. Here results for routes generated by cars and humans are shown on the left and right respectively. The points on the graph represent different values for Retrieval Score Threshold. \vspace{-0.2cm}} 
    \label{fig:AirborneROC}
  \end{figure}

Fig.~\ref{fig:AirborneROC} shows the ROC for the task 12 and 13. Recall that the Retrieval Score Threshold is the minimum path score (generated by Algorithm 1) required in order to declare this path as a correct search result. The points on Fig.~\ref{fig:AirborneROC} represent different Retrieval Score Threshold values for each technique. Those values are equally spaced between the lowest and highest retrieval scores generated by the results. As shown by the generated ROC, our technique outperforms all the other techniques quite noticeably. HDP fails to handle infrequent events. This problem becomes more evident in Task 13 since the beach does not have much activity and hence most queries are considered infrequent.

The accuracy of our tracklets can be quantified by examining Fig.~\ref{fig:AirborneROC}. Here our algorithm (see solid red) shows significant improvement over KLT (see solid black) even though Dynamic programming was used in both methods. For cars the improvement accounts for $105\%$ increase in true positives for a false positive rate of $0.2$. For humans improvement is even higher, with $350\%$ increase in true positives for a false positive rate of $0.2$. The higher improvement for humans is expected as KLT is not robust to low contrast features. To quantify the improvements Dynamic Programming brings to the search problem, we compare our technique against `Tracking Matching' (see Fig.~\ref{fig:AirborneROC}, dashed black). Here our tracklets are used in both methods. However for `Tracking Matching' we declare a tracklet as a match if the mean absolute error between the query track is within a threshold. Fig.~\ref{fig:AirborneROC} shows a significant improvement in true positives by $500\%$ for both humans and cars at a false positive rate of $0.2$ (see red and dashed black). This shows that Dynamic Programming is capable of significantly boosting search results despite any remaining detection errors.

In terms of computational complexity, HDP is the most expensive. Our approach takes an average of 2.9 seconds to process one frame of Task 12 and 4.32 seconds to process one frame of Task 13. Processing time is computed as the average over $2000$ frames. All experiments are performed on an Intel(R) Core(TM) i7-3820 CPU @ 3.60 GHz 3.60GHz with 16.GB RAM. All algorithms are written in an unoptimized MATLAB code except for the KLT extraction technique which is written in C++ by Shi et al. \cite{Shi94}. Note that our storage procedure spares the burden of having to store all the information. For instance, even though we are extracting as much as $4000$ tracklets for task 12, the size of the generated hash table is around $8000$ times smaller than the size of the actual original data (see Table~\ref{tab:task1}). 

  \vspace{-0.1cm}
  \subsection{Discussion}
  \label{sec:discussion}
\vspace{-0.1cm}

  Our method represents a fundamentally different way of approaching the video search problem.  Rather than relying on an abundance of training data or finely-tuned features to differentiate actions of interest from noise, we rely on simple features and causality.  In addition to the clear benefits in terms of a run-time which scales sub-linearly with the length of the video corpus, the simple features and hashing approach render the approach robust to user error as well as poor-quality video. Results demonstrate clearly that causality and temporal structure can be powerful tools to reduce false alarms.
  Another added benefit is how the algorithm scales with query complexity.  In our formulation, the more action components in a query, the more likely it is to differentiate itself from noise.  
  
  There is, of course, non-temporal structure that we have yet to exploit.  Spatial positioning of queries, such as ``The second action component must occur to the northeast of the first one'', or ``The second action component must be near the first one'' is a simple attribute which may further differentiate queries of interest from background noise.  
  
Our method is not void of limitation.  One important limitation is that it requires each activity to be made of discrete states, each of which being describable by a simple feature vocabulary. Complex actions like sign language or actions which are to fast or too small to be identified at the atom level will be difficult to search for.
  
\vspace{-0.1cm}
  \section{Conclusion}
\vspace{-0.1cm}
\label{sec:conclusion}

  We presented a method that summarizes the dynamic content of a surveillance
  video in a way that is compatible with arbitrary user-defined queries.  We
  divide the video into documents each containing a series of atoms.  These
  atoms are grouped together into trees all containing a feature list.  
  We used different features for different types of data. For CCTV footage we used size, color, direction and persistence of
  the moving objects. For Airborne footage we used tracklets of examined objects. Here a viterbi approach for tracklet generation is used. The coordinates of the features trees are then stored into a
  hash-table based feature index.  Hash functions group trees whose content
  is similar.  In this way, search becomes a simple lookup as user-specified
  queries are converted into a table index.  Our method has many advantages.
  First, because of the indexing strategy, our search engine has a complexity of
  $O(1)$ for finding partial matches.  Second, the index requires minimal
  storage. Third, our method requires only local processing, making it suitable
  for facing constant data renewal.  This makes it simple to implement and easy
  to adjust.  Fourth, our method summarizes the entire video, not just the
  predominant modes of activity: it can retrieve any combination of rare,
  abnormal and recurrent activities.

%

\ifCLASSOPTIONcaptionsoff
  \newpage
\fi

%
%
%

\vspace{-0.3cm}
\bibliographystyle{plain}
\bibliography{strings,References}

\begin{thebibliography}{10}

\bibitem{ILIDS}
i-lids.
\newblock \url{computervision.wikia.com/wiki/I-LIDS}.

\bibitem{MIT}
Mit traffic.
\newblock \url{people.csail.mit.edu/xgwang/HBM.html}.

\bibitem{PETS06}
Pets 2006.
\newblock \url{http://ftp.pets.rdg.ac.uk/}.

\bibitem{Abramowitz2002}
M.~Abramowitz and I.~A. Stegun.
\newblock {\em Handbook of Mathematical Functions}.
\newblock Dover, New York, 1970.

\bibitem{Adam08}
A.~Adam, E.~Rivlin, I.~Shimshoni, and D.~Reinitz.
\newblock Robust real-time unusual event detection using multiple
  fixed-location monitors.
\newblock {\em IEEE Trans.\ PAMI}, 30(3):555--560, 2008.

\bibitem{baugh09}
G.~Baugh and A.~Kokaram.
\newblock {A Viterbi tracker for local features }.
\newblock In {\em Proc.\ SPIE Vis. Comm. and Image Process.}, pages 7543--23,
  2009.

\bibitem{Benezeth10}
Y.~Benezeth, P-M. Jodoin, B.~Emile, H.~Laurent, and C.~Rosenberger.
\newblock Comparative study of background subtraction algorithms.
\newblock {\em J. of Elec. Imaging}, 19(3):1--12, 2010.

\bibitem{Horn81}
B.Horn and B.~Schunck.
\newblock Determining optical flow.
\newblock {\em Artif.\ Intell.}, 17(1-3):185--203, 1981.

\bibitem{Bouman98}
C.~Bouman.
\newblock Cluster: An unsupervised algorithm for modeling gaussian mixtures.
\newblock Technical report, Purdue University, 1998.

\bibitem{Calderara07}
S.~Calderara, R.~Cucchiara, and A.~Prati.
\newblock A distributed outdoor video surveillance system for detection of
  abnormal people trajectories.
\newblock In {\em Proc.\ IEEE Conf.\ on Dist.\ Smart Cameras}, pages 364--371,
  2007.

\bibitem{Emonet14}
R.~Emonet, J.~Varadarajan, and J-M Odobez.
\newblock Temporal analysis of motif mixtures using dirichlet processes.
\newblock {\em IEEE Trans.\ PAMI}, 36(1):140--156, 2014.

\bibitem{Gionis99}
A.~Gionis, Piotr Indyk, and R.~Motwani.
\newblock Similarity search in high dimensions via hashing.
\newblock In {\em Proc.\ Int. Conf.\ on Very Large Data Bases}, pages 518--529,
  1999.

\bibitem{Gorelick07}
L.~Gorelick, M.~Blank, E.~Shechtman, M.~Irani, and R.~Basri.
\newblock Actions as space-time shapes.
\newblock {\em IEEE Trans.\ PAMI}, 29(12):2247--2253, 2007.

\bibitem{Hoferlin13}
M.~Hoferlin, B.~Hoferlin, G.~Heidemann, and D.~Weiskopf.
\newblock Interactive schematic summaries for faceted exploration of
  surveillance video.
\newblock {\em IEEE Trans.\ Multimedia}, 15:908--920, 2013.

\bibitem{Jouneau11}
E.~Jouneau and C.~Carincotte.
\newblock Particle-based tracking model for automatic anomaly detection.
\newblock In {\em Proc.\ IEEE Int.\ Conf.\ Image Processing}, pages 513--516,
  2011.

\bibitem{Kuettel10}
D.~Kuettel, M.~Breitenstein, L.~Gool, and V.~Ferrari.
\newblock What's going on? discovering spatio-temporal dependencies in dynamic
  scenes.
\newblock In {\em Proc.\ IEEE Comp. Vis. Pat. Rec.}, pages 1951--1958, 2010.

\bibitem{lee05}
H.~Lee, A.~Smeaton, N.~O'Connor, and N.~Murphy.
\newblock User-interface to a cctv video search system.
\newblock In {\em IEE Int Symp on Imaging for Crime Detec. and Prev.}, pages
  39--43, 2005.

\bibitem{Little13}
S.~Little, K.~Clawson, A.~Mereu, and A.~Rodriguez.
\newblock dentifying and addressing challenges for search and analysis of
  disparate surveillance video archives.
\newblock In {\em In proc of ICICPD}, 2013.

\bibitem{Meessen06}
J.~Meessen, M.~Coulanges, X.~Desurmont, and J-F Delaigle.
\newblock Content-based retrieval of video surveillance scenes.
\newblock In {\em MRCS}, pages 785--792, 2006.

\bibitem{pitie05}
F.~Pitié, S-A. Berrani, R.~Dahyot, and A.~Kokaram.
\newblock Off-line multiple object tracking using candidate selection and the
  viterbi algorithm.
\newblock In {\em Proc.\ IEEE Int.\ Conf.\ Image Processing}, 2005.

\bibitem{Malinici08}
I.~Pruteanu-Malinici and L.~Carin.
\newblock Infinite hidden markov models for unusual-event detection in video.
\newblock {\em IEEE Trans.\ Image Process.}, 17(5):811--821, 2008.

\bibitem{Saligrama10}
V.~Saligrama, J.~Konrad, and P.-M. Jodoin.
\newblock Video anomaly identification: A statistical approach.
\newblock {\em IEEE Signal Process.\ Mag.}, 27:18--33, 2010.

\bibitem{Shechtman07}
E.~Shechtman and M.~Irani.
\newblock Space-time behavior based correlation �or� how to tell if two
  underlying motion fields are similar without computing them?
\newblock {\em IEEE Trans.\ PAMI}, 29(11):2045--2056, 2007.

\bibitem{Shi94}
Jianbo Shi and Carlo Tomasi.
\newblock Good features to track.
\newblock In {\em Proc.\ IEEE Comp. Vis. Pat. Rec.}, pages 593 -- 600, 1994.

\bibitem{Simon09}
C.~Simon, J.~Meessen, and C.~DeVleeschouwer.
\newblock Visual event recognition using decision trees.
\newblock {\em Multimedia Tools and App.}, 2009.

\bibitem{SW81}
T.F. Smith and M.S. Waterman.
\newblock Identification of common molecular subsequences.
\newblock {\em J. of Molecular Biology}, 147:195--197, 1981.

\bibitem{Stringa98}
E.~Stringa and C.~Regazzoni.
\newblock Content-based retrieval and real time detection from video sequences
  acquired by surveillance systems.
\newblock In {\em Proc.\ IEEE Int.\ Conf.\ Image Processing}, pages 138--142,
  1998.

\bibitem{Sujatha11}
C.~Sujatha and U.~Mudenagudi.
\newblock A study on keyframe extraction methods for video summary.
\newblock In {\em in proc of ICCICS}, pages 73--77, 2011.

\bibitem{Thornton11}
J.~Thornton, J.~Baran-Gale, D.~Butler, M.~Chan, and H.~Zwahlen.
\newblock Person attribute search for large-area video surveillance.
\newblock In {\em IEEE Int. Conf. on Tech. for Homeland Security.}, pages
  55--61, 2011.

\bibitem{Tian08}
Y-L. Tian, A.~Hampapur, L.~Brown, R.~Feris, M.~Lu, A.~Senior, C-F. Shu, and
  Y.~Zhai.
\newblock Event detection, query, and retrieval for video surveillance.
\newblock In Zongmin Ma, editor, {\em Artificial Intelligence for Maximizing
  Content Based Image Retrieval}. Information Science Reference, 2008.

\bibitem{Veeraraghavan07}
H.~Veeraraghavan, N.~Papanikolopoulos, and P.~Schrater.
\newblock Learning dynamic event descriptions in image sequences.
\newblock In {\em Proc.\ IEEE Comp. Vis. Pat. Rec.}, pages 1--6, 2007.

\bibitem{Wang09}
X.~Wang, X.~Ma, and E.Grimson.
\newblock Unsupervised activity perception in crowded and complicated scenes
  using hierarchical bayesian models.
\newblock {\em IEEE Trans.\ PAMI}, 31(3):539--555, 2009.

\bibitem{Xiang08}
T.~Xiang and S.~Gong.
\newblock Video behavior profiling for anomaly detection.
\newblock {\em IEEE Trans.\ PAMI}, 30(5):893--908, 2008.

\bibitem{Yang09}
Y.~Yang, B.~Lovell, and F.~Dadgostar.
\newblock Content-based video retrieval (cbvr) system for cctv surveillance
  videos.
\newblock In {\em Proc of Dig. Img. Comp. Tech. and App.}, pages 183--187,
  2009.

\bibitem{Yeo08}
C.~Yeo, P.~Ahammad, K.~Ramchandran, and S.~Sastr.
\newblock High speed action recognition and localization in compressed domain
  videos.
\newblock {\em IEEE Trans.\ Circuits Syst.\ Video Technol.}, 18(8):1006 --
  1015, 2008.

\bibitem{Yilmaz08}
A.~Yilmaz and M.~Shah.
\newblock A differential geometric approach to representing the human actions.
\newblock {\em Comput.\ Vis.\ Image Und.}, 109(3):335--351, 2008.

\bibitem{Zhu05}
X.~Zhu, X.~Wu, A.~Elmagarmid, Z.~Feng, and L.~Wu.
\newblock Video data mining:semantic indexing and event detection from the
  association perspective.
\newblock {\em IEEE Trans.\ Know\ Data En}, 17:665--677, 2005.

\end{thebibliography}

%




\end{document}